\documentclass{article}

\usepackage{PRIMEarxiv}

\usepackage[utf8]{inputenc} % allow utf-8 input
\usepackage[T1]{fontenc}    % use 8-bit T1 fonts
\usepackage{hyperref}       % hyperlinks
\usepackage{url}            % simple URL typesetting
\usepackage{booktabs}       % professional-quality tables
\usepackage{amsfonts}       % blackboard math symbols
\usepackage{amsmath}
\usepackage{natbib}
\usepackage{multirow}
\usepackage{nicefrac}       % compact symbols for 1/2, etc.
\usepackage{microtype}      % microtypography
\usepackage{lipsum}
\usepackage{fancyhdr}       % header
\usepackage{graphicx}       % graphics
\usepackage{subfig}
\usepackage{makecell}
\graphicspath{{media/}}     % organize your images and other figures under media/ folder

\title{Can human clinical rationales improve the performance and explainability of clinical text classification models?
}

\author{
  Christoph Metzner \\
  The University of Tennessee \\
  Knoxville, TN, USA\\
  \texttt{cmetzner@vols.utk.edu} \\
  \And
  Shang Gao and Drahomira Herrmannova \\
  Thomson Reuters \\
  Alpharetta, GA, USA\\
  \texttt{\{shang.gao,dasha.herrmannova\}@thomsonreuters.com} \\
   \And
  Heidi A. Hanson\\
  Oak Ridge National Laboratory\\
  Oak Ridge, TN, USA\\
  \texttt{hahanson@ornl.gov} \\
}

\begin{document}
\maketitle

\begin{abstract}

AI-driven clinical text classification is vital for explainable automated retrieval of population-level health information. This work investigates whether human-based clinical rationales can serve as additional supervision to improve both performance and explainability of transformer-based models that automatically encode clinical documents. We analyzed 99,125 human-based clinical rationales that provide plausible explanations for primary cancer site diagnoses, using them as additional training samples alongside 128,649 electronic pathology reports to evaluate transformer-based models for extracting primary cancer sites. We also investigated sufficiency as a way to measure rationale quality for pre-selecting rationales. Our results showed that clinical rationales as additional training data can improve model performance in high-resource scenarios but produce inconsistent behavior when resources are limited. Using sufficiency as an automatic metric to preselect rationales also leads to inconsistent results. Importantly, models trained on rationales were consistently outperformed by models trained on additional reports instead. This suggests that clinical rationales don't consistently improve model performance and are outperformed by simply using more reports. Therefore, if the goal is optimizing accuracy, annotation efforts should focus on labeling more reports rather than creating rationales. However, if explainability is the priority, training models on rationale-supplemented data may help them better identify rationale-like features. We conclude that using clinical rationales as additional training data results in smaller performance improvements and only slightly better explainability (measured as average token-level rationale coverage) compared to training on additional reports.

\end{abstract}

% keywords can be removed

\section{Introduction}
AI-driven clinical support systems are revolutionizing patient care by enhancing a clinician's access to detailed patient information and evidence-based recommendations \cite{elhaddad2024ai}. These systems rely largely on clinical information extracted from an exponentially growing body of unstructured electronic health records (e.g., electronic pathology reports), necessitating efficient automated extraction through machine learning (ML)-based text classification models. To gain the trust of patients and clinicians in high-stakes scenarios such as patient care, these ML models must demonstrate both high performance and explainability. In addition to ongoing algorithmic advances, recent work in non-clinical domains utilizes human-based rationales as an additional supervision signal to improve model performance and explainability \cite{wiegreffe2021teach}. Although initial results are promising, it remains unclear if human-based \textit{clinical} rationales will enable similar model improvements in clinical text classification tasks, including the automated encoding of clinical documents. Clinical documents are particularly challenging to process owing to their use of complex clinical language (e.g., jargon, abbreviations, varying terminology) and multiple contextual topics (e.g., patient history information, institutional details, diagnostic information), both of which exacerbate the difficulties associated with learning from human-based clinical rationales (hereafter referred to as \textit{rationales}). To this end, this work investigates how clinical text classification models are impacted by incorporating rationales in the model training. 

Rationales are concise, plausible explanations that adequately justify a document's ground-truth label, such as a clinical diagnosis \cite{carton2020evaluating, wiegreffe2021teach}. Rationales are either (i) abstractive free-form text created by humans or generative models \cite{chan2023knife, yu2023explanation, kwon2024large, savage2024diagnostic} or (ii) extractive text highlights retrieved by a human \cite{zaidan2007using, zhang2016rationale, bao2018deriving, strout2019human, zhong2019fine, arous2021marta, zhang2021human} or a model \cite{sharma2018learning, taylor2021rationale, zhang2021explain, pruthi2022evaluating}. Although some research has explored the use of rationales to enhance model explainability \cite{strout2019human, arous2021marta, zhang2021explain, pruthi2022evaluating, gurrapu2023rationalization, resck2024exploring}, the primary application of rationales lies in improving the performance of ML models \cite{zaidan2007using, zhang2016rationale, bao2018deriving, sharma2018learning, arous2021marta, taylor2021rationale, zhang2021explain, resck2024exploring}. Generally, researchers leverage rationales as a form of prior domain knowledge in three key ways: (i)~as supplementary samples that enable models to discover new connections between input features and output labels \cite{zaidan2007using, sharma2018learning, resck2024exploring}, (ii)~as fine-grained supervision to provide additional ground-truth labels at the word or sentence level \cite{zhang2016rationale, arous2021marta, taylor2021rationale}, and (iii)~as guidance for internal model components (e.g., attention mechanisms) to direct a model's focus toward rationale-like features \cite{bao2018deriving, zhong2019fine, arous2021marta, zhang2021human, pruthi2022evaluating}. However, more recent work has explored large language models (LLMs) for generating abstractive rationales to help the model to self-rationalize its output \cite{chan2023knife, yu2023explanation, kwon2024large, savage2024diagnostic}. 

While previous work on rationales in the clinical domain explored generative approaches \cite{taylor2021rationale, kwon2024large, savage2024diagnostic}, our study focuses on extractive rationales, highlighted rationales retrieved verbatim from a given cancer pathology report, to enhance specialized discriminative models for the automated encoding of electronic pathology reports with primary cancer site diagnoses. Interestingly, our preliminary experiments showed that these rationale highlights used as additional token-level supervision, either as ground-truth labels for token-sequence classification models or as attention masks, resulted in lower performance compared to utilizing them as supplementary independent training samples (\autoref{tab:intro_preliminary_results}). These findings, coupled with previous research demonstrating the effectiveness of rationales as supplementary training input \cite{zaidan2007using, sharma2018learning} or as the sole training input \cite{resck2024exploring}, motivated us to explore the potential of our rationales as supplementary training data. The main idea of this approach is that rationales represent information pruned of noise, thereby providing a direct link to a given primary cancer site \cite{jacovi2021contrastive} and allowing the model to learn more robust relationships between rationale-like input and the label space and thus improve explainability and performance.

\section{Objective} % DONE 
This study examines how incorporating rationales for primary cancer site diagnoses as an additional supervision signal during model training can enhance transformer-based models for automated clinical document classification. Given the resource-intensive nature of clinical text data annotation, we investigate these enhancements focusing on the classification of electronic pathology reports with the primary cancer site diagnoses collected by the National Cancer Institute's (NCI's) Surveillance, Epidemiology, and End Results (SEER) program from 2011 to 2022 \cite{NCI-SEER}. Our investigation has three main objectives: (i)~to evaluate whether including rationales as additional training data can improve model performance and explainability, contrasted against models trained without additional rationale training data and models trained with additional report training data; (ii)~to understand how the impact of rationale supplementation differs between high-resource scenarios, where rationales are available for all classes, and low-resource scenarios, where rationales exist for only a few classes; and (iii)~to assess whether filtering rationales using the \textit{sufficiency} metric \cite{deyoung2019eraser, carton2020evaluating} affects model performance. The contributions of this work are as follows:

\begin{itemize}
    \item We are the first to investigate extractive, human-based clinical rationales for improving clinical text classification models.
    \item Our results show that supplementing the training data with human-based clinical rationales can improve model performance and explainability in high-resource scenarios. 
    \item Our findings demonstrate that models trained solely on human-based clinical rationales are outperformed by those trained on full and complementary information, highlighting the complexity of electronic pathology reports.
    \item We show that the sufficiency metric is unreliable for improving model performance when used to quantify rationale quality.
    \item Based on our results, we recommend that annotation efforts focus on labeling new, additional documents to maximize model performance.
\end{itemize}

\section{Materials and Methods} %154
% word count: 895
\subsection{Data}

\subsubsection{SEER Cancer Pathology Reports} % DONE

We obtained a total of 128,649 electronic pathology reports from the New Jersey, Louisiana, and Utah cancer registries of the NCI's SEER program (\autoref{tab:dataset_statistics}). Each report is associated with a distinct Cancer-Tumor-Case (CTC) ID; each CTC may be linked to multiple reports. Following annotation standards of the North American Association of Central Cancer Registries, all reports were manually annotated by certified tumor registrars (CTRs) with ground-truth labels for multiple critical cancer data elements, including primary cancer site, subsite, laterality, histology, and behavior. However, this study examines the classification of 69 primary cancer sites annotated across all 128,649 reports, exploring the performance impact of integrating rationales in the training of clinical text classification models. A subset of the reports has been annotated by CTRs with rationales that explain the selected primary cancer site diagnosis.
 
\subsubsection{Generation of Human-Based Clinical Rationales}  \label{mm_rationales_data} 

The selection of cancer pathology reports for rationale annotation followed a systematic workflow established by the Surveillance, Epidemiology, and End Results Program Data Management System (SEER*DMS) for autocoding pathology reports with essential cancer data elements (i.e., site, laterality, histology, and behavior) using an ML-based API \cite{hsu2024machine}. In particular, the rationale annotations were created during the quality control step of succesfully autocoded reports (those where all four cancer data elements were automatically extracted with sufficient confidence, i.e., met a predefined minimum prediction probability threshold). During this quality control step, approximately 10\% of the autocoded reports were manually reviewed by data oncology specialists. If a specialist disagrees with the top-ranked diagnosis suggested by the API for a given report, then they were tasked to provide the correct diagnosis along with supporting highlights from the report, i.e., rationale(s), or a comment in the absence of the information relevant for the provided diagnosis. In our study, we focused only on highlights related to the primary cancer site that matched at least one instance of the rationale in the associated report, which resulted in a total of 89,718 reports with 99,125 rationales (\autoref{tab:dataset_statistics}). Notably, a single report may have been annotated with multiple rationales if a data oncology specialist saw fit; see Appendix \autoref{tab:app_distribution_rationales_training}. During model training, we incorporated each rationale as an independent sample to supplement the training dataset, similar to previous approaches \cite{zaidan2007using, sharma2018learning, resck2024exploring}; note that we also trained models on only rationales for a specific analysis. The models had no access to rationales during inference. Additional details on the rationales, their generation, and the class frequency distributions are available in Appendix~\ref{app:data} and \cite{SEERDMS2020, hsu2024machine}.

\subsubsection{Dataset Partitioning}

We included nearly all reports with at least one rationale in the training split, reserving a small fraction for the testing split so that we could perform model explainability analyses. To populate the validation and testing splits, reports were randomly sampled from the available total population of SEER electronic pathology reports; these reports do not possess any rationales. We ensured that reports associated with a distinct CTC were confined to a single split and that class distributions were consistent across all splits. Lastly, to provide a control to our rationale-trained models, we randomly selected an additional set of pathology reports ensuring no overlap in CTCs and matching the same class distribution as the set of rationales. This control training dataset enables us to investigate whether annotations efforts should be spent on retrieving rationales or labeling new, unlabeled reports.

\begin{table}[h!]   
    \centering

\caption{Descriptive statistics of the SEER cancer pathology reports highlighted with human-created rationales. Each report in the training split is associated with at least one rationale. The test split consists of 19,566 non-annotated reports and 2,446 reports annotated with a single rationale.}
    \resizebox{\linewidth}{!}{%
    \begin{tabular}{ c c c c c c c c c}
    \specialrule{1.0pt}{0pt}{0pt}
    & \multicolumn{3}{c}{Reports} & \multicolumn{5}{c}{Rationales} \\
    \cmidrule(lr){2-4}
    \cmidrule(lr){5-9}
     &  & \multicolumn{2}{c}{Mean Text Lengths (\textbf{s})} & &  \multicolumn{2}{c}{Mean Text Lengths (\textbf{s})} & \multicolumn{2}{c}{Coverage in \% (\textbf{s})}\\
     \multirow{-3}{*}{Split} & \multirow{-2}{*}{N} & Word & Subword & \multirow{-2}{*}{N} & Word & Subword & Word & Subword \\
        \specialrule{1.0pt}{0pt}{0pt}

     Train & 87,252 & 594.4 (523.3) & 1029.5 (897.0) & 96,679 & 13.6 (16.8) & 26.9 (29.4) & 4.1 (5.5) & 4.3 (5.4) \\
     Val & 19,385 & 540.5 (466.2) & 946.5 (817.3) & 0 & -- & -- & -- & -- \\
     Test & 22,012 & 548.8 (485.2) & 961.0 (848.1) & 2,446 & 12.7 (15.9) & 25.7 (28.3)  & 3.8 (5.1) & 4.1 (5.3) \\
     Total & 128,649 & 578.5 (509.2) & 1005.3 (877.9) & 99,125 & 13.6 (16.7) & 26.9 (29.4) & 4.1 (5.5) & 4.3 (5.4) \\
     \specialrule{1.0pt}{0pt}{0pt}
     \multicolumn{9}{p{\linewidth}}{The maximum rationale length was set to 128 word tokens (i.e., $mean + 2\times std$); longer rationales were removed. Maximum sequence length was set to 4,096 subword tokens.}\\

    \specialrule{1.0pt}{0pt}{0pt}
    \end{tabular}
    }  % CLOSING BRACKET FOR resizebox
    \label{tab:dataset_statistics}
\end{table}

\subsection{Model Architectures} % DONE

While recent work on advanced models like LLMs demonstrated promising performance in supervised classification of breast cancer pathology reports (n < 1000) \cite{sushil2024comparative}, several factors led us to select the simpler clinical longformer (CLF) as the base text-encoder architecture for all evaluated models: (i) unlike LLMs, CLF's open transformer-based architecture provides easy access to internal model parameters and allows architectural modifications; (ii) its simpler architecture enables efficient fine-tuning at lower computational costs compared to the repeated fine-tuning of LLMs; and (iii) the CLF was pretrained on medical text demonstrating strong performances on clinical text classification tasks \cite{li2022clinical, metzner2024attention} supporting \cite{sushil2024comparative} observation that simpler model's may achieve performance comparable to LLMs when trained on large annotated datasets.

We focused on two primary models: the baseline version of the CLF (CLF-BS) and the CLF extended with the recently published deformable phrase-level attention mechanism (CLF-DPLA) \cite{metzner2024deformable}. We framed the task as a multiclass classification problem in which a model parameterized by $\theta$ aims to map a given input text sequence $x$---report, rationale, or complement---to its corresponding class label $y$ (i.e., primary cancer site). A detailed description of the model architectures and the classification process can be found in Appendix~\ref{app:mm:model_architecture}. 

\subsection{Model Evaluation} % DONE

We evaluated the performance of all models by using quantitative metrics established in the multiclass text classification literature and measured the performance accuracy and the macro-averaged F1-score \cite{kowsari2019text, gao2021limitations, minaee2021deep}. Accuracy measures a model's overall performance, while the macro-averaged F1-score provides class-wise performance evaluation by giving equal weight to each class regardless of its frequency in the dataset \cite{santos2022automatic}. We used the widely accepted normal approximation \cite{raschka2018model} to compute 95\% confidence intervals (see Appendix \ref{app:model_training}). 

\subsection{Evaluation of Rationales} 

Rationales can be useful for enhancing model performance and explainability \cite{strout2019human}. Explainability of ML models is generally decomposed into two components: (i) faithfulness (i.e., how well the explanation describes the inner mechanisms of a model for a given prediction) and (ii) plausibility (i.e., how well a human understands the provided explanation). Because our rationales are generated by subject matter experts, we assume that they are all highly plausible explanations for a given primary cancer site diagnosis. However, despite their plausibility, rationales may not be faithful. Therefore, it is important to evaluate rationale faithfulness. Previous work evaluated faithfulness by using two automatic metrics, \textit{sufficiency} and \textit{comprehensiveness} \cite{carton2020evaluating,yu2019rethinking, deyoung2019eraser}, which measure a rationale's ability to reproduce the same prediction as the full report and the degree to which a rationale contains all the information relevant to the prediction, respectively. We used the implementations of both metrics proposed by \cite{carton2020evaluating}. 

Sufficiency (\autoref{eq:sufficiency}) measures how a rationale enables the model to reproduce the same prediction as the full report by comparing the prediction probabilities of the full report versus only the rationale:

\begin{equation} \label{eq:sufficiency}
    \text{Suff}(x, \hat{y}, \alpha) = 1 - max(0, p(\hat{y}|x) - p(\hat{y}|x, \alpha)),
\end{equation}

where x represents a full report, $\hat{y}=argmax[p(y|x)]$, and $\alpha$ is a binary mask indicating whether a word belongs to a rationale associated with x. In contrast, comprehensiveness (\autoref{eq:comprehensiveness}) measures how well a rationale encapsulates the information relevant for making the correct prediction by comparing the prediction probabilities between the full report and the complement of the rationale.

\begin{equation} \label{eq:comprehensiveness}
    \text{Comp}(x, \hat{y}, \alpha) = max(0, p(\hat{y}|x) - p(\hat{y}|x, 1 - \alpha))
\end{equation}

A highly comprehensive rationale should contain more relevant information for the correct prediction than its complement and result in more accurate predictions. The literature argues that a rationale is faithful when it has both high sufficiency and comprehensiveness. Finally, we investigate whether sufficiency is suitable as a metric to discriminate between lower- and higher-quality rationales to improve model training and performance. We experimented with a sufficiency threshold of 0.2, which removes the least sufficient rationales, and a threshold of 0.817 represents the 90\% percentile of rationales with false predictions for falsely predicted reports; removing roughly 88\% of the rationales that led to false predictions in preliminary experiments.

\section{Results} 
% word count: 874
\subsection{Model Performance Across Varying Training Regimens}
% 250
\autoref{tab:res_gen_rat_improve} summarizes the classification accuracies and macro-averaged F1-scores for the CLF-BS and CLF-DPLA trained on varying training data regimens: (i)~reports only, (ii)~reports supplemented with the full set of available rationales, (iii)~reports supplemented with a subset of sufficient rationales for the two sufficiency thresholds of 0.2 and 0.817, and (iv)~reports supplemented with 96,679 additional labeled SEER electronic pathology reports as a control experiment. The results for both evaluated models show that training on reports supplemented with rationales (regimens ii and iii) can improve performance over training solely on reports (regimen i). However, the control experiments showed that training models on more reports with full information (regimen iv) outperformed the baseline (regimen i) and rationale-based models (regimens~ii and~iii) across both metrics. The CLF-DPLA architecture consistently outperformed the CLF-BS across all training regimens and metrics.

Both models benefited from supplementing the training dataset (regimen i) with either all rationales (regimen~ii) or a set of sufficient rationales (regimen~iii), but the CLF-BS did not meaningfully improve in accuracy or macro-averaged F1-score when trained on (regimen ii). However, when trained on sufficient rationales (regimen iii), the macro F1-score did increase by 3\%. In contrast, the CLF-DPLA achieved significant performance improvements when trained on (ii) versus (i). The achievable performance improvements for the CLF-DPLA trained on sufficient rationales (regimen iii) are inconsistent---the accuracy score remains unchanged, the macro-averaged score for the sufficiency threshold of 0.2 slightly increases, and the macro-averaged score for the threshold of 0.817 decreases.  
\begin{table}[h!]
    \caption{Comparison of accuracy and F1-macro scores for varying training regimen for the CLF-BS and the CLF extended with the deformable phrase-level attention mechanism (CLF-DPLA) when performing the classification of the primary cancer site in SEER cancer pathology reports. Both models were trained on only reports ($n_{reports}=87,252$), reports supplemented with human-based clinical rationales ($n_{rationales}=96,679$), reports supplemented with sufficient rationales, or reports supplemented with additional reports ($n_{reports, additional}=96,679$) as a control. All models were evaluated on the same test dataset $n_{test} = 22,012$. The set of additional reports follows the same distribution as the full set of rationales. The $^\ast$ indicates significant performance improvements over reports only. The best performance per model is in bold.}  
    \vspace{1mm}
    \centering
    \resizebox{\linewidth}{!}{%
    \begin{tabular}{c c c c c c c}
    \specialrule{1pt}{0.5pt}{0.5pt}
     &  
     & \multicolumn{2}{c}{\shortstack{Number of Added\\Rationales}} 
     & \multicolumn{2}{c}{\shortstack{Number of Added Sufficient\\Rationales (>Threshold)}}
     & \shortstack{Number of\\Additional Reports} \\
    \cmidrule(lr){3-4} \cmidrule(lr){5-6} \cmidrule(lr){7-7}
    \multirow{-2}{*}{Model} & \multirow{-2}{*}{Metric} & 0 & 96,679 & 94,176 (> 0.2) & 80,950 (> 0.817) &96,679 \\
    \specialrule{1pt}{0.5pt}{0.5pt}
    & \multirow{-2}{*}{Accuracy}
    & \shortstack{0.885\\(0.881, 0.889)} 
    & \shortstack{0.888\\(0.883, 0.892)} 
    & \shortstack{0.890\\(0.886, 0.894)} 
    & \shortstack{0.888\\(0.884, 0.892)} 
    & \shortstack{\textbf{0.903$^\ast$}\\\textbf{(0.899, 0.907)}} \\
    \multirow{-3}{*}{CLF-BS} & \multirow{-2}{*}{F1-Macro} 
    & \shortstack{0.567\\(0.561, 0.574)}
    & \shortstack{0.571\\(0.564, 0.578)}  
    & \shortstack{0.586$^\ast$\\(0.580, 0.593)} 
    & \shortstack{0.580\\(0.570, 0.593)} 
    & \shortstack{\textbf{0.611$^\ast$}\\\textbf{(0.604, 0.617)}} \\
    \specialrule{0.75pt}{0.5pt}{0.5pt}
    & \multirow{-2}{*}{Accuracy}
    & \shortstack{0.900\\(0.896, 0.904)} 
    & \shortstack{0.904\\(0.900, 0.908)}
    & \shortstack{0.904\\(0.900, 0.908)}
    & \shortstack{0.904\\(0.900, 0.908)}
    & \shortstack{\textbf{0.908$^\ast$}\\\textbf{(0.904, 0.912)})} \\
    \multirow{-3}{*}{CLF-DPLA} & \multirow{-2}{*}{F1-Macro} 
    & \shortstack{0.597\\(0.590, 0.603)}
    & \shortstack{0.617$^\ast$\\(0.610, 0.623)} 
    & \shortstack{0.620$^\ast$\\(0.614, 0.626)} 
    & \shortstack{0.610\\(0.603, 0.616)}
    & \shortstack{\textbf{0.625$^\ast$}\\\textbf{(0.619, 0.631)}}  \\
    \specialrule{1pt}{0.5pt}{0.5pt}
    \multicolumn{7}{p{\linewidth}}{Sufficiency of $>$0.2 achieved the best performance improvements across all tested thresholds.
    Sufficiency of $>$0.817 represents the 90\% percentile of rationales with false predictions for falsely predicted reports.
    The 95\% confidence interval is computed via normal approximation \cite{raschka2018model}.} \\
    \specialrule{1pt}{0.5pt}{0.5pt}
    \end{tabular}
    } % closing resizebox bracket
    \label{tab:res_gen_rat_improve}
\end{table}

\subsection{Class-Wise Model Performance in Low-Resource Scenario for 10 Common Primary Cancer Sites}
%215

To study how integrating rationales impacts class-wise model performance in low-resource scenarios, we simulated these scenarios by adding (to the training reports) a small, randomly selected subset of $m\in[100; 500; 1,000]$ rationales for 10 common primary cancer sites. This enabled us to test the potential benefit of adding rationales when data is scarce. From the results in \autoref{tab:classwise_sufficiency_performance}, we can surmise that (i)~the addition of rationales may lead to improved class-wise performance, (ii)~sufficient rationales do not show consistent positive impacts on the class-wise performance, (iii)~the number of already available reports for a given class in the training dataset (infrequent vs. frequent classes) has no consistent impact on the achievable class-wise performance, and (iv) the class-wise performance improvements do not follow a consistent pattern (e.g., multiple classes experience similar good performance with only 100 and 1,000 added rationales while having a reduced performance with 500 additional rationales), see kidney (C64). To provide a full picture of the performance of our tested models in low-resource scenarios with limited availability of rationales, we also report their overall performance in \autoref{tab:performance_results_m_suff_rationales} in \autoref{app:results}. Generally, adding a few rationales reduces overall performance for both models. This trend persists in scenarios in which the training data is augmented with sufficient rationales. 

\begin{table}[h!]
    \caption{Performance improvements of the CLF-BS and CLF-DPLA models when classifying primary cancer sites from SEER pathology reports. Class-wise F1-scores for 10 common cancer sites are shown as a function of $m\in[100, 500, 1,000]$ additional rationales supplemented to the original reports dataset (n=87,252). HRS is hematopoietic and reticuloendothelial systems.}
    \vspace{1mm}
    \centering
    \resizebox{\textwidth}{!}{%

    \begin{tabular}{c c c c c c c c c c c c c c c c c}
    \specialrule{2pt}{1pt}{1pt}
        \multicolumn{2}{l}{ICD-O-3 Primary Cancer Site}  
         & \multicolumn{3}{c}{Stomach (C16)} 
         & \multicolumn{3}{c}{Colon (C18)} 
         & \multicolumn{3}{c}{Pancreas (C25)} 
         & \multicolumn{3}{c}{HRS (C42)} 
         & \multicolumn{3}{c}{Skin (C44)}         \\
\cmidrule(lr){1-2}
\cmidrule(lr){3-5}\cmidrule(lr){6-8}\cmidrule(lr){9-11}\cmidrule(lr){12-14}\cmidrule(lr){15-17}
\multicolumn{2}{l}{Number of Training Reports}
&\multicolumn{3}{c}{1,612}
&\multicolumn{3}{c}{3,679}
&\multicolumn{3}{c}{1,849}
&\multicolumn{3}{c}{7,636}
&\multicolumn{3}{c}{5,646} \\
\cmidrule(lr){1-2}\cmidrule(lr){3-5}\cmidrule(lr){6-8}\cmidrule(lr){9-11}\cmidrule(lr){12-14}\cmidrule(lr){15-17}

\multicolumn{2}{l}{Number of Training Rationales}

& 1,728 & 1,617 & 1,275
& 4,145 & 4,044 & 3,164
& 2,085 & 1,905 & 1,459
& 8,122 & 8,089 & 7,418
& 6,349 & 6,204 & 5,670 \\

\multicolumn{2}{l}{Ratio of Training Rationales}
& 1.0 & 0.94 & 0.74
& 1.0 & 0.98 & 0.76 
& 1.0 & 0.91 & 0.70
& 1.0 & 1.0 & 0.91 
& 1.0 & 0.98 & 0.89\\

\cmidrule(lr){1-2}\cmidrule(lr){3-5}\cmidrule(lr){6-8}\cmidrule(lr){9-11}\cmidrule(lr){12-14}\cmidrule(lr){15-17}
 & & \multicolumn{3}{c}{Sufficiency}& \multicolumn{3}{c}{Sufficiency}& \multicolumn{3}{c}
 {Sufficiency}& \multicolumn{3}{c}{Sufficiency}& \multicolumn{3}{c}{Sufficiency} \\
\multirow{-2}{*}{Model} & \multirow{-2}{*}{\shortstack{Number of Added\\Rationales}} 
     & > 0 & > 0.2 & > 0.817 
     & > 0 & > 0.2 & > 0.817 
     & > 0 & > 0.2 & > 0.817 
     & > 0 & > 0.2 & > 0.817 
     & > 0 & > 0.2 & > 0.817 \\
    \specialrule{1.5pt}{0.5pt}{0.5pt}
    & 0
    & 0.789 & -- &--
    & 0.847 & -- &--
    & 0.802 & --&--
    & 0.904 & --&--
    & 0.961 & --&-- \\
    \cmidrule(lr){2-17}
    & 100 
    & 0.787 & 0.782 & 0.770 % C16
    & 0.845 & 0.853 & 0.854 % C18
    & 0.800 & 0.808 & 0.803 % C25
    & 0.909 & 0.900 &  0.908 % C42
    & 0.957 & 0.961 & 0.963 \\ % C44
    \cmidrule(lr){2-17}
    & 500 
    & 0.781 & 0.772 & 0.783 % C16
    & 0.847 & 0.847 & 0.851 % C18
    & 0.807 & 0.808 & 0.811 % C25
    & 0.904 & 0.892 & 0.900 % C42
    & 0.964 & 0.955 & 0.962 \\ % C44
    \cmidrule(lr){2-17}
    \multirow{-5}{*}{CLF-BS} & 1,000$^\ast$
    & 0.766 & 0.772 & 0.773 % C16
    & 0.854 & 0.853 & 0.854 % C18
    & 0.812 & 0.816 & 0.801 % C25
    & 0.898 & 0.901  & 0.901 % C42
    & 0.959 & 0.962 & 0.962 \\ % C44
    \specialrule{1pt}{0.5pt}{0.5pt}
    & 0
    & 0.813 & --&--
    & 0.898 & --&--
    & 0.826 & --&--
    & 0.908 & --&--
    & 0.971 & --&-- \\
    \cmidrule(lr){2-17}
    & 100 
    & 0.788 & 0.798 & 0.804  % C16
    & 0.893 & 0.891 & 0.894 % C18
    & 0.827 & 0.816 & 0.822 % C25
    & 0.907 & 0.908 & 0.911 % C42
    & 0.971 & 0.971 & 0.974 \\% C44
    \cmidrule(lr){2-17}
    & 500 
    & 0.804 & 0.796 & 0.797  % C16
    & 0.896 & 0.877 & 0.888 % C18
    & 0.824 & 0.836 & 0.834  % C25
    & 0.906 & 0.909 & 0.910 % C42
    & 0.973 & 0.970 & 0.974 \\ % C44
    \cmidrule(lr){2-17}
    \multirow{-5}{*}{CLF-DPLA}& 1,000$^\ast$ 
    & 0.795 & 0.806 & 0.781 % C16
    & 0.889 & 0.887 & 0.894 % C18
    & 0.812 & 0.827 & 0.819  % C25
    & 0.902 & 0.901 & 0.903 % C42
    & 0.969 & 0.969 & 0.970\\ % C44
    \specialrule{2pt}{1pt}{1pt}
\multicolumn{2}{l}{ICD-O-3 Primary Cancer Site} 
& \multicolumn{3}{c}{Breast (C50)} 
         & \multicolumn{3}{c}{Corpus uteri (C54)} 
         & \multicolumn{3}{c}{Prostate gland (C61)} 
         & \multicolumn{3}{c}{Kidney (C64)} 
         & \multicolumn{3}{c}{Thyroid gland (C73)} \\
\cmidrule(lr){1-2}
\cmidrule(lr){3-5}\cmidrule(lr){6-8}\cmidrule(lr){9-11}\cmidrule(lr){12-14}\cmidrule(lr){15-17}
\multicolumn{2}{l}{Number of Training Reports}
&\multicolumn{3}{c}{25,160}
&\multicolumn{3}{c}{3,521}
&\multicolumn{3}{c}{1,847}
&\multicolumn{3}{c}{1,068}
&\multicolumn{3}{c}{1,500} \\
\cmidrule(lr){1-2}\cmidrule(lr){3-5}\cmidrule(lr){6-8}\cmidrule(lr){9-11}\cmidrule(lr){12-14}\cmidrule(lr){15-17}

\multicolumn{2}{l}{Number of Training Rationales}

& 28,408 & 28,196 & 27.017 
& 3,790 & 3,615 & 2,925
& 1,964 & 1,865 & 1,682 
& 1,134 & 1,060 & 973
& 1,573 & 1,509 & 1,455 \\

\multicolumn{2}{l}{Ratio of Training Rationales}
& 1.0 & 0.99 & 0.95
& 1.0 & 0.95 & 0.77 
& 1.0 & 0.95 & 0.86
& 1.0 & 0.93 & 0.86 
& 1.0 & 0.96 & 0.92 \\

\cmidrule(lr){1-2}\cmidrule(lr){3-5}\cmidrule(lr){6-8}\cmidrule(lr){9-11}\cmidrule(lr){12-14}\cmidrule(lr){15-17}

 & & \multicolumn{3}{c}{Sufficiency}& \multicolumn{3}{c}{Sufficiency}& \multicolumn{3}{c}{Sufficiency}& \multicolumn{3}{c}{Sufficiency}& \multicolumn{3}{c}{Sufficiency} \\
\multirow{-2}{*}{Model} & \multirow{-2}{*}{\shortstack{Number of Added\\Rationales}} 
     & > 0 & > 0.2 & > 0.817 
     & > 0 & > 0.2 & > 0.817 
     & > 0 & > 0.2 & > 0.817 
     & > 0 & > 0.2 & > 0.817 
     & > 0 & > 0.2 & > 0.817 \\
    \specialrule{1.5pt}{0.5pt}{0.5pt}
    & 0
    & 0.982 & --&--
    & 0.907 & --&--
    & 0.816 & --&--
    & 0.521 & --&--
    & 0 & -- &-- \\
    \cmidrule(lr){2-17}
    & 100
    & 0.980 & 0.980 & 0.980  % C50
    & 0.907 & 0.900 & 0.907  % C54
    & 0.829 & 0.818 & 0.806 % C61
    & 0.522 & 0.518 & 0.479 % C61
    & 0.226 & 0.048 & 0.175 \\ % C73 
    \cmidrule(lr){2-17}
    & 500
    & 0.981 & 0.983 & 0.981 % C50
    & 0.909 & 0.902 & 0.910 % C54
    & 0.845 & 0.830 & 0.833 % C61
    & 0.554 & 0.553 & 0.553 % C61
    & 0.222 & 0.165 & 0.044 \\ % C73 
    \cmidrule(lr){2-17}
    \multirow{-5}{*}{CLF-BS} & 1,000$^\ast$
    & 0.982 &0.981 & 0.979 % C50
    & 0.904 & 0.901 & 0.899  % C54
    & 0.836 & 0.836 &  0.831 % C61
    & 0.432 &0.541  & 0.539  % C61
    & 0.089 & 0.217 &  0.044 \\ % C73 
    \specialrule{1pt}{0.5pt}{0.5pt}
    & 0
    & 0.984 & --&--
    & 0.934 & --&--
    & 0.868 & --&--
    & 0.511 & --&--
    & 0.307 & --&-- \\
    \cmidrule(lr){2-17}
    & 100
    & 0.986 & 0.986 & 0.985 % C50
    & 0.927 & 0.927 & 0.928 % C54
    & 0.864 & 0.855 & 0.854 % C61
    & 0.477 & 0.570 & 0.564 % C61
    & 0.180 & 0.039 & 0.122 \\ % C73 
    \cmidrule(lr){2-17}
    & 500
    & 0.986 & 0.987 & 0.987 % C50
    & 0.927 & 0.918 & 0.922 % C54
    & 0.841 & 0.861 & 0.876  % C61
    & 0.503 & 0.504 & 0.537 % C61
    & 0.313 & 0.276 & 0.377 \\ % C73 
    \cmidrule(lr){2-17}
    \multirow{-5}{*}{CLF-DPLA}& 1,000$^\ast$ 
    & 0.987 & 0.987 & 0.984 % C50
    & 0.922 & 0.927 & 0.928  % C54
    & 0.871 & 0.859 & 0.855 % C61
    & 0.521 & 0.575 & 0.542 % C61
    & 0.152 & 0.140 & 0.089 \\ % C73 
    \specialrule{2pt}{1pt}{1pt}
    \end{tabular}
    }
    \label{tab:classwise_sufficiency_performance}
\end{table}

\subsection{Human-Based Clinical Rationales and Model Explainability}

Next, we analyze the potential impact of rationales on a model's explainability from the perspective of attention. For this analysis, we focused on the better performing CLF-DPLA. Because the DPLA deploys two target attention mechanisms, we can use the generated attention scores\footnote{We acknowledge the controversy and discussion about the suitability of attention scores for providing sufficient model explainability \cite{bibal2022attention}.} as a gauge for the relative importance of a token at the token and phrase levels for the primary cancer site classification task. For all 2,446 rationale-annotated test reports, we calculated the overlap ratio between the rationale tokens and the tokens with the highest attention scores by using the 90, 95, and 98 attention score percentiles as thresholds. Notably, because transformers utilize a byte-pair encoded vocabulary with subwords and we want to represent the ratios at the word level, we recreated the word-level attention scores as the average of all subword-level scores. We hypothesize that the type of training input impacts the token-level rationale coverage ratios, so we computed the ratios for models trained on only rationales, only complements, only reports, reports supplemented with all rationales, and reports supplemented with additional labeled reports.

\begin{figure}
    \centering
    \includegraphics[width=1\linewidth]{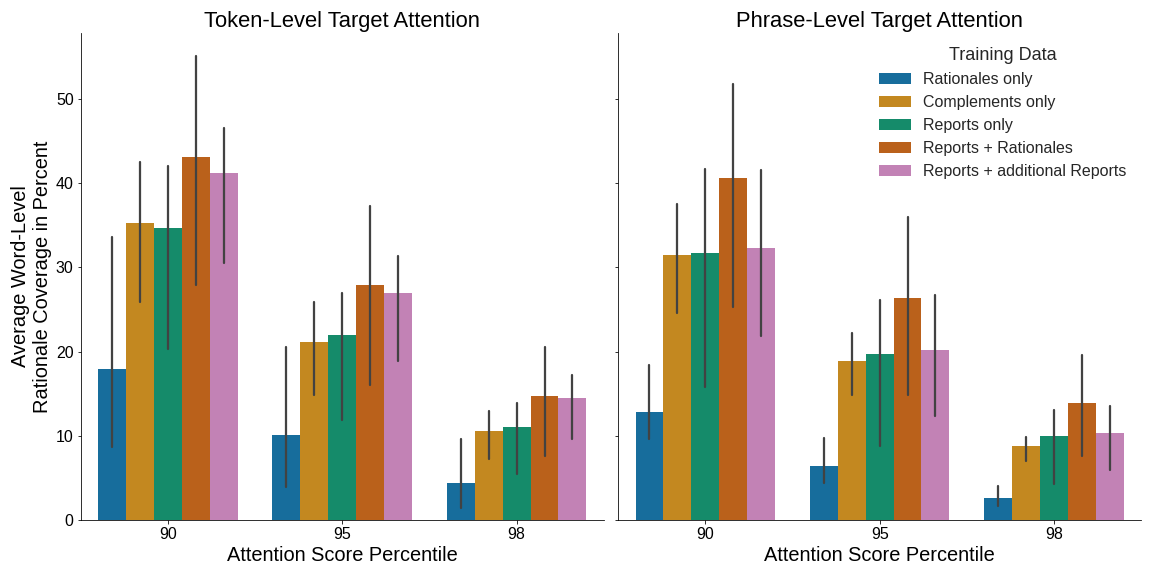}
    \caption[Average token-level rationale coverage for the CLF-DPLA.]{Average token-level rationale coverage across different attention score percentiles for DPLA's word- and phrase-level target attention mechanisms. Rationale coverage is calculated as the proportion of tokens deemed highly important (based on attention scores) that overlap with human-annotated rationales. Results are aggregated from 2,446 test electronic pathology reports. The error bars represent the standard deviation between the average token-level rationale coverage ratios across different random initialization seeds.} 
    \label{fig:rationale_coverage_img1}
\end{figure}

\autoref{fig:rationale_coverage_img1} presents the token-level rationale coverage ratio averaged across all the test documents for the CLF-DPLA and indicates that the ratios are impacted by the selected attention score percentile and the type of training data. For both attention mechanisms, models trained on rationales achieved the lowest overlap, whereas models trained on reports and rationales achieved the highest overlap with the available rationale annotations. Interestingly, supplementing the training data with additional reports enables models to achieve similar high overlap with rationales for the token-level attention mechanism but only medium overlap for the phrase-level target attention mechanism. In contrast, models trained on only reports or complements had a medium overlap with the rationale annotations associated with the test documents. Appendix \ref{app:explainability} provides a more detailed breakdown of the average token-level rationale coverage ratios by prediction (i.e., correct versus false), rationale length, and training regimen--specific overlap of the highlighted tokens. In short, the rationale overlap (i.e., average token-level rationale coverage) is higher for correct predictions across all training regimens, tend to follow a unimodal distribution across sequence length with a peak coverage for sequences ranging from 6 to 15 words, and the different training regimens highlight up to 60\% of the same tokens associated with the rationales.  

\section{Discussion}
% 1800
\subsection{Impact of Human Expert--Derived Rationales on Model Performance}
We explored the impact of rationales on clinical text classification models used for automated medical encoding of electronic pathology reports with primary cancer site diagnoses. Our experiments revealed that rationales can positively influence model performance. However, we observed decreased overall and inconsistent class-wise performance in low-resource scenarios, a finding that suggests potential issues with rationale quality. Interestingly, these inconsistencies in model performance remain for models trained on sufficient rationales. Lastly, the increase in classification performance from augmenting the training data with rationales was smaller than the increase from our control model that simply trained on more pathology reports. 

Although our findings align with previous research on enhancing ML model performance using rationales \cite{arous2021marta, zhang2021explain, resck2024exploring} in high-resource scenarios, the observed improvements in accuracy are somewhat underwhelming in terms of the annotation effort required to achieve performance increases given the extensive addition of rationales---effectively doubling the size of the training dataset. This is particularly notable when compared to the minimal improvements over the baseline performance (i.e., models trained on the reports without rationales) and the lower performance compared with the control models trained on additional labeled reports. Therefore, we think that the rather modest improvements achieved with supplementing rationales over the models trained on only reports and their low performance compared with models trained on a larger volume of pathology reports raises concerns about the return on investment for annotating rationales. Essentially, when improving the model accuracy is the goal, we recommend allocating resources to acquiring and annotating new, unlabeled samples instead of annotating rationales in already collected and labeled documents. Notably, we agree with \cite{zaidan2007using} observation that the simultaneous annotation of the ground-truth label (i.e., diagnosis) and the rationale maximizes the return on investment during data annotation. 

Given the considerable effort required to annotate rationales and the low availability of datasets with a large number of annotated rationales \cite{wiegreffe2021teach}, it is important to evaluate model performance in low-resource scenarios. Contrary to our expectations, the addition of a few rationales---randomly sampled from 10 common primary cancer sites---to the training data caused a drop in overall model performance and an interesting display of inconsistent trends in class-wise performance. The drop and inconsistent trends in performance may be caused by unwanted class-wise interactions between the introduced rationales, general quality-related issues with the set of available rationales, or rationales introducing signals that possibly contradict the information a model finds useful in the original training reports for a given class; this information may stem from spurious correlations or artifacts introduced through the data \cite{gardner2021competency}. Furthermore, the mismatch in signal between rationales and reports has been described by \cite{resck2024exploring} as data distribution shift possibly causing a decrease in model performance. This data distribution shift could also explain the inconsistent model performances in high- versus low-resource scenarios, in which models may experience the introduction of rationales as noise in low-resource scenarios because the class-wise input features provided by the rationales differ from the features provided by the reports of the same class. Although the analysis of possible interactions between classes is worthwhile, it is beyond the scope of this study. 

Another approach to explain the mismatch in signal between rationales and reports and the potential quality-related issues can be borrowed from the explainable AI literature. In this space, rationales are seen as explanations \cite{strout2019human}, which can be faithful (i.e., able to explain a model's inner decision making) and/or plausible (i.e., understandable by humans). Although we think our rationales provide highly plausible explanations because they were derived by subject matter experts---acknowledging typical variations in annotation quality---our rationales may not necessarily provide faithful information.

In our first analysis, we evaluated the faithfulness of our rationales by comparing the model performance when trained on only rationales, on full reports, or on complements. According to the literature, faithful rationales should ideally enable models to perform comparably to those trained on full reports \cite{lyu2024towards}. However, our experiments found that models trained on only rationales achieved the poorest performance of the group, whereas report-trained models achieved the highest performance (\autoref{tab:comparison_reports_comps_rationales}). These findings suggest that (i)~ML models rely on a broad range of information presented in clinical documents to optimally learn the decision boundary space; (ii)~plausible explanations (i.e., text that a human would find crucial for a clinical diagnosis) may not necessarily be as important to a ML model as they are to humans; (iii)~ML models may find seemingly irrelevant information, possible spurious correlations, or data artifacts to be critical in developing their decision boundary space; (iv)~cancer pathology reports are complex and contain multiple crucial concepts that a model potentially relies on during its decision-making; and (v)~to a model, rationales represent only a single source of crucial information provided by the cancer pathology reports. Additionally, rationales appear to provide essential information in reports associated with minority classes indicated by the reduced F1-macro scores of models trained on complements compared with models trained on reports. Our findings align with the assessment of Carton et al., who argued that models are susceptible to biases through artifacts in the data \cite{carton2020evaluating}, thereby causing inconsistent or false evaluation of rationales, and who empirically observed that human rationales may not provide sufficient information for the model to exploit for prediction \cite{carton2022learn}. Lastly, we are not surprised that models trained on only rationales performed far worse than their report- and complement-trained counterparts given that our rationales (on average) cover only roughly 4\% of the full text and many rationales consist of only a few words.

\begin{table}[h!]
    \centering
        \caption{Comparison of accuracy and F1-macro scores for different training input types for the base CLF-BS and the CLF-DPLA when performing the classification of the primary cancer site in SEER cancer pathology reports trained on only the reports ($n=87,252$), rationales ($n=96,679$), or the rationale complements ($n=96,679$). The models were tested on the training split containing the reports, validation ($n=19,385$), and test splits ($n=22,012$).}
    \label{tab:comparison_reports_comps_rationales}
    \vspace{1mm}
    \resizebox{0.6\linewidth}{!}{%
    \begin{tabular}{c c c c c c}
    \specialrule{1pt}{0.5pt}{0.5pt}
    & & & \multicolumn{3}{c}{Training Data Input} \\
    \cmidrule{4-6}
    \multirow{-2}{*}{Model} & \multirow{-2}{*}{Split} & \multirow{-2}{*}{Metric} & Rationales & Complements & Reports \\
    \specialrule{1pt}{0.5pt}{0.5pt}

    & & Accuracy & 0.840&0.916 & 0.932   \\
    &\multirow{-2}{*}{Train} & F1-Macro 
    & 0.468 
    & 0.644 
    & 0.707 \\
    \cmidrule{2-6}
    & & Accuracy 
    & 0.802 
    & 0.878 
    & 0.880 \\
    &\multirow{-2}{*}{Val} & F1-Macro 
    & 0.437 
    & 0.537  
    & 0.552\\
        \cmidrule{2-6}

    & & Accuracy 
    & 0.801 
    & 0.881 
    & 0.885 \\
    \multirow{-6}{*}{CLF-BS}&\multirow{-2}{*}{Test} & F1-Macro 
    & 0.437 
    & 0.553 
    & 0.567 \\
    
    \specialrule{1pt}{0.5pt}{0.5pt}  
    & & Accuracy 
    & 0.691 
    & 0.916 
    & 0.937 \\
    &\multirow{-2}{*}{Train} & F1-Macro 
    & 0.328 
    & 0.670 
    & 0.755 \\
        \cmidrule{2-6}

    & & Accuracy 
    & 0.793 
    & 0.900 
    & 0.901 \\
    &\multirow{-2}{*}{Val} & F1-Macro 
    & 0.433 
    & 0.583 
    & 0.600 \\
        \cmidrule{2-6}

    & & Accuracy 
    & 0.780 
    & 0.900  
    & 0.900\\
    \multirow{-6}{*}{CLF-DPLA}&\multirow{-2}{*}{Test} & F1-Macro 
    & 0.426 
    & 0.579 
    & 0.597 \\
    \specialrule{1pt}{0.5pt}{0.5pt}  

    \end{tabular}
}
\end{table}

\begin{figure}
    \centering
    \includegraphics[width=\linewidth]{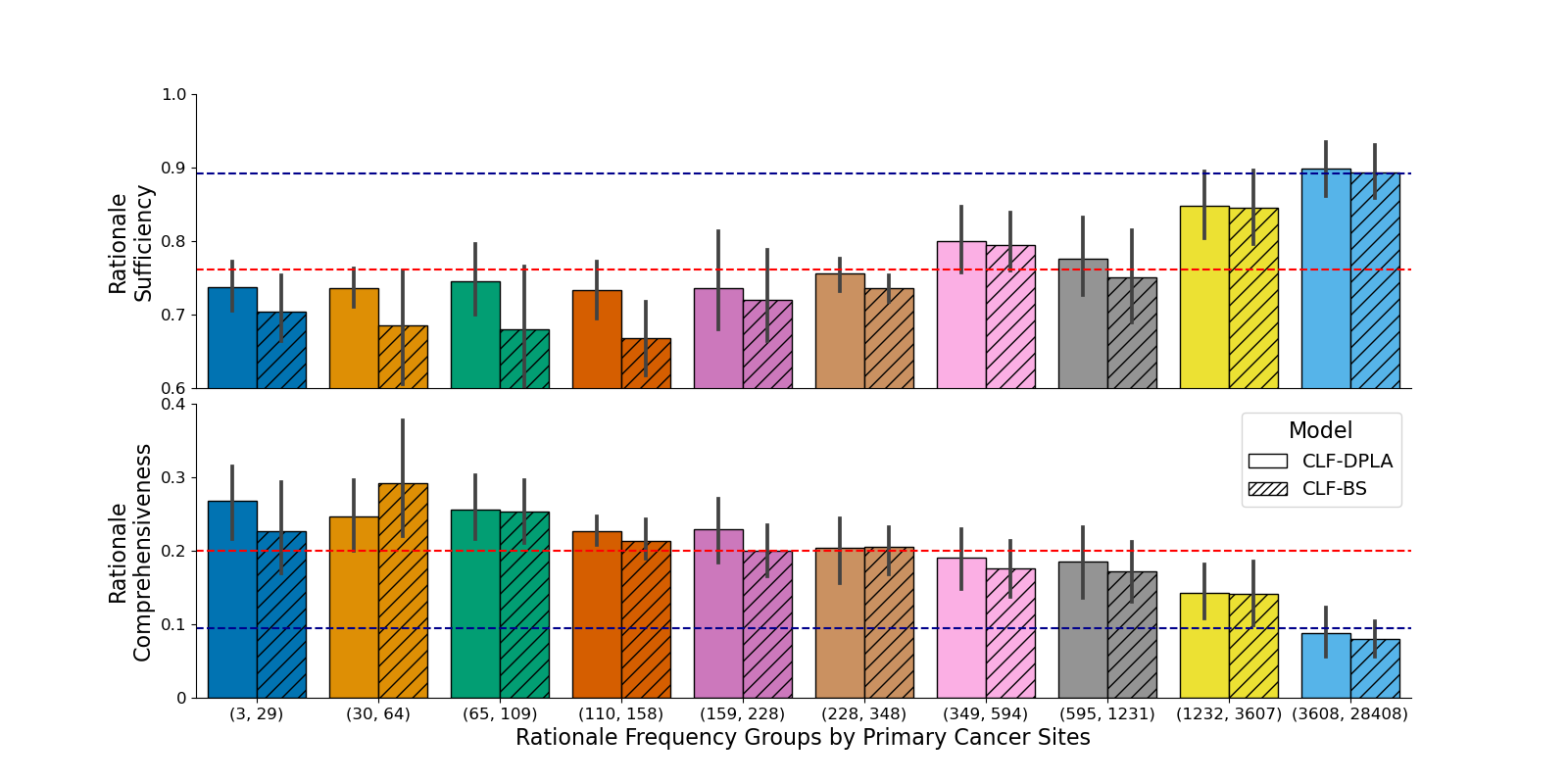}
    \caption{Evaluation of rationale faithfulness decomposed into sufficiency and comprehensiveness for the CLF-BS and CLF-DPLA models for all evaluated primary cancer sites grouped by their frequency occurrence. The blue-dashed line depicts the mean score averaged across all samples, and the red-dashed line is the mean score averaged across all classes.}
    \label{fig:sufficiency_comprehensiveness}
\end{figure}

Our second analysis followed recent work that evaluated rationale quality by decomposing faithfulness into two key properties: sufficiency and comprehensiveness \cite{deyoung2019eraser, carton2020evaluating}. Sufficiency attempts to measure how well a rationale can substitute the full information to reproduce the prediction, whereas comprehensiveness assesses how well a rationale encapsulates all information relevant for the prediction. Therefore, a high-quality faithful rationale should have both high sufficiency and high comprehensiveness \cite{carton2020evaluating}. Furthermore, in their comprehensive survey on datasets with annotated rationales, Wiegreffe et al. recommended avoiding the use of low-sufficiency rationales and to assess whether the collected rationales truly explain the ground-truth label (i.e., primary cancer site) \cite{wiegreffe2021teach}. We computed the sufficiency and comprehensiveness scores by using the prediction probabilities generated from the experiments that compared model training on rationales, full reports, and complements. The scores are shown in \autoref{fig:sufficiency_comprehensiveness}.

Generally, our rationales achieve sufficiency scores similar to those in prior published work on non-clinical, human-based rationales. However, the comprehensiveness scores for our rationales are lower than the published ones \cite{carton2020evaluating}. The low comprehensiveness scores further support the claim made above: the rationales, although containing crucial information for a given report's diagnosis, only cover a small fraction of the relevant information in the full report. 

Interestingly, both metrics are impacted by the support for each class in the dataset, with sufficiency increasing and comprehensiveness decreasing with an increased number of samples for a given class. It is unclear from our set of experiments whether this trend for both metrics is caused by the actual quality associated with the rationales across classes or by having more class-wise support, thereby allowing the ML-based model to become more confident in its predictions. Given that both metrics are based on prediction probabilities, we conclude that the latter case is more likely, especially considering that each rationale was created for a specific cancer pathology report and primary cancer site diagnosis. Although minority classes occur infrequently, we do not anticipate that label frequency will significantly impact the rationale annotation process or substantially degrade rationale quality.

Finally as a third analysis, we wondered if the sufficiency metric could be used as a preselection threshold to determine higher-quality rationales to improve model performance (see Tables~\ref{tab:classwise_sufficiency_performance} and \ref{tab:performance_results_m_suff_rationales}). Generally, preselecting rationales based on sufficiency scores does not consistently improve performance when tested across the full set of rationales in low-resource scenarios. Although we see improved overall performance when augmenting the training dataset with \textit{sufficient} rationales, the performance in low-resource scenarios is inconsistent, with performance dropping for colon (C18) and for thyroid gland (C73), indicating that sufficiency alone is not an ideal metric for rationale preselection.

\subsection{Impact of Human Expert--Derived Rationales on Model Explainability}
% word count: 465
Next, we discuss the potential impact of our rationales on model explainability. Explainability is understood as a model's ability to provide plausible explanations for its predictions (i.e., the greater the overlap between features highlighted by the model as important and human annotated rationales, the more plausible the model \cite{wiegreffe2021teach}). Model plausibility is especially important in medicine. Similar to \cite{resck2024exploring}, we hypothesized that supplementing the training regimen with plausible explanations in the form of our rationales can teach a model to focus its attention on more rationale-like features. We computed the average token-level rationale coverage for 2,446 test documents using attention scores generated by the CLF-DPLA when trained on reports supplemented with rationales. We then contrasted those results with reports only, rationales only, complements only, and reports supplemented with additional reports (\autoref{fig:rationale_coverage_img1}). 

Our analysis shows that the type of information used for model training influences rationale coverage, and this indicates that model plausibility can be impacted by the selection of the training data. The results further indicate that rationales alone are not a sufficient source of information to \textit{teach} models plausibility. In particular, models do not necessarily need the rationale information, and models trained on complements or full reports achieve similar average token-level rationale coverage. The analysis shows that the different types of training input result in an overlap of highlighted rationales of up to 60\% (see Appendix~\ref{app:int_overlap}). It is only when we combine rationales with full reports that rationales are helpful for teaching models to look for rationale-like features. Furthermore, our results also show that statistically similar levels of plausibility can be achieved through supplementing the training data with additional reports.

Although we think that rationales that encapsulate plausible explanations for a primary cancer site diagnosis can be used to teach models to identify rationale-like features as important, we strongly encourage annotators to ensure that the extracted rationales contain sufficient information from the report without introducing noise through the addition of words not relevant to the actual diagnosis. Appendix \ref{app:int_length} shows that the average token-level rationale coverage is impacted by rationale length. Although our experiments show that rationales may not consistently improve model performance, rationales can positively impact model explainability. Our findings suggest that clinical rationales serve as an effective source of information for priming AI models to generate plausible explanations while maintaining robust performance. This balance between explainability and accuracy is crucial for the successful implementation of AI-driven clinical support systems. Therefore, model optimizations must focus on both aspects to enhance the trustworthiness of such systems and increase their acceptance among medical professionals \cite{zhang2021explain, resck2024exploring}. Interestingly, our experiments showed that this trade-off exists, as highlighted by the improved rationale coverage and relatively consistent performance for models trained on rationales.

\section{Conclusion, Limitations, and Future Work}

% word count 379
We investigated the impact of integrating human-based clinical rationales into training data to improve the performance of transformer-based clinical text classification models used to classify SEER electronic pathology reports with the primary cancer site diagnosis. Our results show that adding rationales can improve model performance in high-resource scenarios but achieves inconsistent performance in low-resource scenarios. Additionally, rationale-trained models are outperformed by models trained on a dataset augmented with more reports. Our results highlight the importance of measuring and ensuring rationale faithfulness with appropriate automatic metrics because any quality-related issues can degrade model performance.

The main conclusions of this work are as follows: (i)~If improving model performance is the goal, then the highest return on investment for data annotations is achieved by annotating additional documents instead of extracting rationales from already available reports. (ii)~For scenarios in which additional reports with full information are unavailable, the annotation of all available reports with rationales can effectively improve a model's overall performance. However, the selective annotation for frequent classes may not consistently improve class-wise performance; the selective annotation of minority classes should be considered. (iii)~Utilizing automatic metrics (e.g., sufficiency) to assess a rationale's quality leads to inconsistent improvements in model performance. (iv)~Finally, including rationales in the training data can improve model explainability by priming a model to pay more attention to rationale-like subsequences in a document during inference. Please note that we only investigated rationales as additional training data and the impact of rationales on model performance depends on their implementation method and may differ from ours. 

Although this study offers valuable insights for using rationales to improve clinical text classification models, one must consider its limitations and potential avenues for future research. A major limitation of this study is the introduction of a sampling bias that occurs when collecting data from the reports annotated with rationales. This sampling bias is expressed by creating rationales for reports that failed the initial quality control and a registry-specific data origin check. Another limitation is that our study focuses primarily on the evaluation of extractive rationales to improve discriminative models and neglects abstractive rationales created by generative models. Future research could explore the strengths of LLMs to either create \textit{synthetic} extractive rationales or investigate methods that force LLMs to self-rationalize when classifying pathology reports with the primary cancer sites. Another direction could be the investigation of automatic metrics that appropriately measure the quality of rationales. Lastly, future work could investigate learning curves of adding rationales to better understand the optimal ratio of rationales to reports needed for effective model training.

\section*{Acknowledgments}

The authors would like to acknowledge the contribution to this study from other staff in the participating central cancer registries. These registries are supported by the National Cancer Institute’s Surveillance, Epidemiology, and End Results (SEER) Program, the Centers for Disease Control and Prevention’s National Program of Cancer Registries (NPCR), and/or state agencies, universities, and cancer centers.  The participating central cancer registries include the following: New Jersey working under contract numbers SEER: 75N91021D0000/75N91021F00001 and NPCR: NU58DP006279; Louisiana working under contract numbers SEER: HHSN2612018000071/HHSN26100002 and NPCR: NU58DP0063; and Utah working under contract numbers SEER: HHSN261201800016I and NPCR: NU58DP007131. 

We would like to acknowledge Chad A. Melton and Patrycja Krawcyuk with the Oak Ridge National Laboratory due to their assistance with querying the data during the review process.

\section*{IRB Statement}
The study on the SEER-based electronic pathology reports was approved under the IRB DOE000152.

\section*{Funding Statement}
This work was supported by resources of the Oak Ridge Leadership Computing Facility at the Oak Ridge National Laboratory, which is supported by the US Department of Energy's Office of Science under Contract No. DE-AC05-00OR22725. This work was produced by UT-Battelle, LCC, under Contract No. DE-ACO5-000R22725 with the U.S. Department of Energy. Publisher acknowledges the U.S. Government license to provide public access under the DOE Public Access Plan (http://energy.gov/downloads/doe-public-access-plan).

\section*{Competing Interests Statement}

The authors declare no competing interests.

\section*{Contributorship Statement}

Christoph Metzner designed, implemented, and executed the study and analyzed the results of all experiments. Shang Gao, Drahomira Herrmannova, and Heidi Hanson helped designing and analyzing the experiments and results. Christoph Metzner wrote the manuscript, Shang Gao, Drahomira Herrmannova, and Heidi Hanson reviewed and edited the manuscript.

\clearpage

% Bibliography
%Bibliography
\bibliographystyle{unsrt}  
\bibliography{references.bib}

%%%%%%%%%%%%%% APPENDIX %%%%%%%%%%%%%%%%%
\clearpage

% \blankpage

\appendix
\addcontentsline{toc}{section}{Appendices}

\renewcommand\thepage{\thesection.\arabic{page}}
\setcounter{page}{1}

\renewcommand\thetable{\thesection.\arabic{table}}
\setcounter{table}{0}

\renewcommand\theequation{\thesection.\arabic{equation}}
\setcounter{equation}{0}

% this was formerly called introduction

\section{Materials and Methods}
\setcounter{table}{0}
\setcounter{equation}{0}
\counterwithin{figure}{section}
\setcounter{figure}{0}
\setcounter{page}{1}

\subsection{Data} \label{app:data}

\begin{figure}[h!]
  \centering 
  \subfloat[][\centering Flow chart of rationale selection process.]{\includegraphics[width=0.4\linewidth]{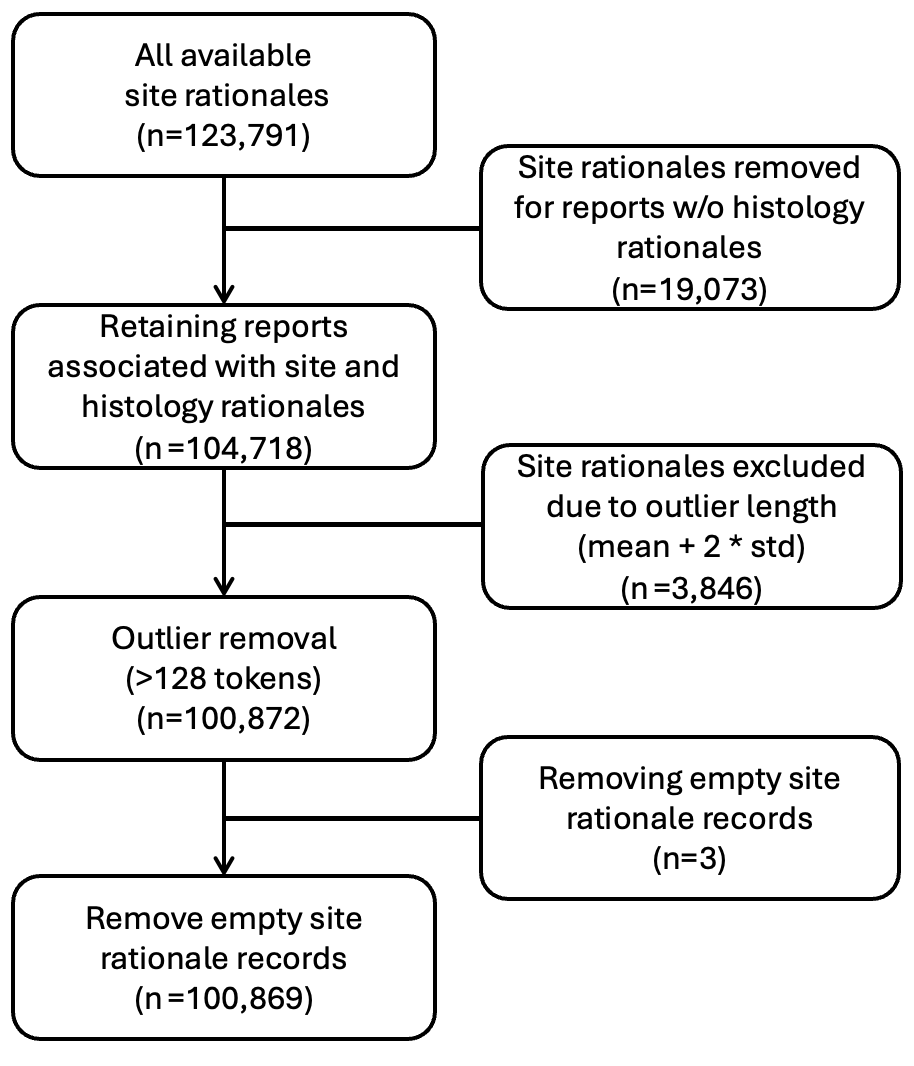}}% 
  \qquad 
  \subfloat[][\centering Flow chart of rationale selection process with matching instances in their associated reports.]{\includegraphics[width=1\linewidth]{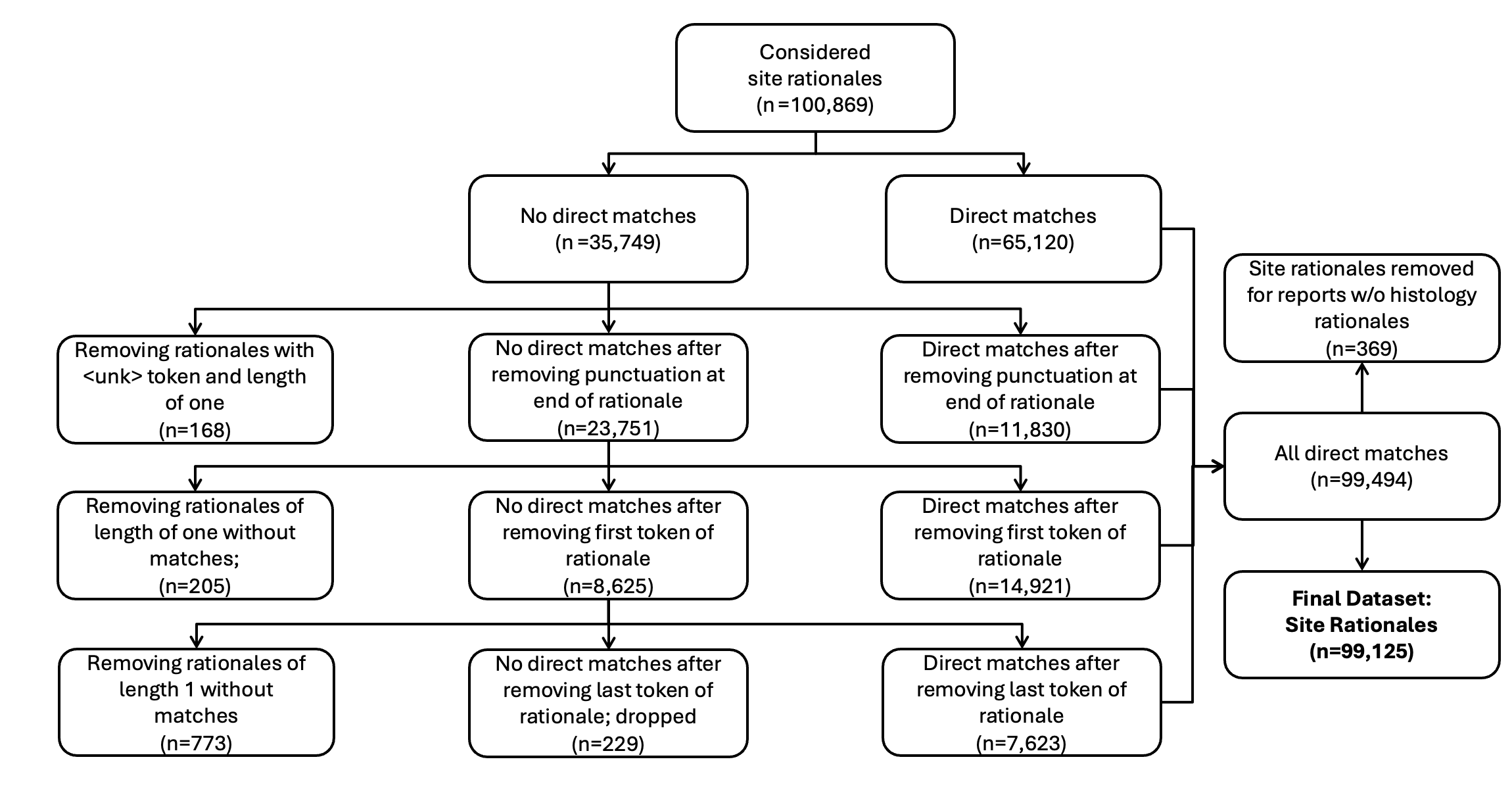}}% 
  \vspace{1mm}
  \caption{Overview of the primary cancer site rationale selection and matching process. Initially, only reports with site and histology rationales were considered. However, the histology rationales were omitted owing to a change in research priority.}
  \label{app:fig:rationale_selection_matching_charts}
\end{figure}

This work considers human-based clinical rationales that provide plausible explanations for the primary cancer site assigned to electronic pathology reports. The included flowchart provides a detailed overview of generating the final rationale dataset (\autoref{app:fig:rationale_selection_matching_charts}). Notably, although we initially considered analyzing records annotated with rationales for the primary cancer site \textit{and} the histological type, we ultimately changed our research priority to focus on only primary cancer sites, which led us to not include rationales associated with histology. Because the experiments were already underway when we changed our strategy, we did not utilize the full set of available primary cancer site rationales, and we acknowledge a potential introduction of sampling bias when choosing our subset of primary cancer site rationales.

% https://deainfo.nci.nih.gov/advisory/fac/0222/Lowy.pdf this is the source for the 17% (from 2022)
\subsubsection{Generation of Human-Based Clinical Rationales}

The selection of cancer pathology reports for rationale annotation followed a systematic workflow established by the Surveillance, Epidemiology, and End Results Program Data Management System (SEER*DMS) for autocoding pathology reports with essential cancer data elements (site, laterality, histology, and behavior) using an ML-based API \cite{hsu2024machine}. In particular, the rationale annotations were created during the quality control step of successfully autocoded reports (those where all four cancer data elements were automatically extracted with sufficient confidence, i.e., met a predefined minimum prediction probability threshold). During this quality control step, approximately 10\% of the autocoded reports were manually reviewed by data oncology specialists. If a specialist disagrees with the top-ranked diagnosis suggested by the API for a given report, then they were tasked to provide the correct diagnosis along with a supporting highlight from the report, i.e., rationale, or a comment in the absence of the information relevant for the provided diagnosis. More details on the workflow can be found in \cite{hsu2024machine} and on the rationale annotation in \cite{SEERDMS2020}.

In our study, we focus only on highlights related to the primary cancer site that match at least one instance of the rationale in the associated report, which results in a total of 89,718 reports with 99,125 rationales. We exclude all reports annotated with comments because they may not necessarily be extractive. Notably, the exact spatial location of these highlights was lost during preprocessing because the spatial location indicated the positions (i.e., row and column position in a text file) of the first and last characters of a given highlight in the raw text file. Therefore, we considered all occurrences of a given highlight as a rationale within the report. We treat our rationales as independent samples and integrate the rationale information into the model training as additional training data, which is similar to the approaches of previous work \cite{zaidan2007using, sharma2018learning, resck2024exploring}. Additional details on the generation of rationales can be found in \cite{SEERDMS2020}. 

We included nearly all reports with at least one rationale in the training split and reserved a small fraction for the testing split to enable us to perform model explainability analyses (\autoref{tab:dataset_statistics}). Additionally, models that perform the automated medical encoding of clinical documents do not have access to rationales during inference. The number of distinct rationales per document ranges from one to seven, with approximately 99\% of the training reports having either one or two rationales. Each report in the test set has one rationale, allowing us to maximize the number of available rationales in the training dataset by avoiding the removal of reports with multiple rationales. For each rationale, we created binary masks that indicate whether a token is part of a rationale by matching the rationale with a substring of the report. Although the majority of the 65,120 rationales were matched directly, a total of 34,005 rationales were matched after further cleaning, and a total of 1,744 rationales were dropped because they were not matchable. See Appendix~\ref{app:data} for a detailed overview of the rationale generation. Because the initial, raw set of rationales contained very long sequences, longer than 1,000 tokens with a maximum of 1,235 tokens, we assumed those greater than the average rationale length plus 2$\times$ the standard deviation---maximum rationale length is 128 tokens---were erroneous, and we excluded these from our analysis.

We used the binary masks to create complements for the rationales (i.e., complement samples include the full pathology report minus the information contained by a rationale). Similar to \cite{carton2020evaluating}, we drop all rationale tokens from the reports to prevent the model from potentially learning positional information from the rationales. To populate the validation and testing splits, documents were randomly sampled from the available total population of SEER cancer pathology reports (these reports do not possess any rationales). We ensured that reports associated with a distinct Cancer-Tumor-Case (CTC) were confined to a single split and that class distributions were consistent across all splits. Lastly, we randomly selected an additional set of cancer pathology reports, ensuring no overlap in CTCs and matching the same distribution as the set of rationales, to investigate whether annotations efforts should be spent on retrieving rationales or labeling new, unlabeled reports.

\autoref{tab:distribution_data} shows the distribution of records, cases, and patients by registry and data split. Because our training dataset contains documents annotated with multiple rationales, we show the breakdown in \autoref{tab:app_distribution_rationales_training}. Each test report was only annotated with a single rationale. \autoref{tab:rationale_length} shows a breakdown of the number of rationales associated with a distinct range of rationale lengths. \autoref{tab:freq_counts} and Table A.4 (cont.) show the frequency count of the primary cancer site across the training, validation, and test splits containing SEER cancer pathology reports. We also show the frequency count for the primary cancer site for the population of SEER reports annotated with the primary cancer site rationales and the entire population of SEER reports used to populate our validation and test data splits.

\begin{table}[htb]
    \centering
    \caption{Overview of SEER electronic pathology reports.}
    \vspace{1mm}
    \begin{tabular}{c c c c | c}
         \specialrule{1.0pt}{0.5pt}{0.5pt}
         Registry & Train & Val & Test & Total \\
         \specialrule{1.0pt}{0.5pt}{0.5pt}
         New Jersey & 72,079 & 8,773 & 10,826 & 91,678 \\
         Louisiana & 13,135 & 7,484 & 7,843 & 28,462 \\
         Utah & 2,038 & 3,128 & 3,343 & 8,509 \\
         \specialrule{0.75pt}{0.5pt}{0.5pt}
         Reports & 87,252 & 19,385 & 22,012 & 128,649 \\
         Tumor cases & 51,136 & 7,916 & 10,351 & 69,403 \\
         Patients & 49,616 & 7,910 & 10,341 & 67,867 \\
         \specialrule{1.0pt}{0.5pt}{0.5pt}
    \end{tabular}
    \label{tab:distribution_data}
\end{table}

\begin{table}[h!]
    \centering
    \caption{Frequency distribution of human-created rationales for SEER cancer pathology reports contained in the training dataset.}
    \vspace{1mm}
    \label{tab:app_distribution_rationales_training}
    \begin{tabular}{c c c}
    \specialrule{1pt}{0.5pt}{0.5pt}
         \shortstack{Number of Rationales\\per Document } & \shortstack{Number of\\Documents} & \shortstack{Total Number of\\Rationales} \\
         \specialrule{1pt}{0.5pt}{0.5pt}
         1 & 78,801  & 78,801 \\
         2 & 7,608& 15,216  \\
         3 & 736 & 2,208 \\
         4 & 92& 368  \\
         5-7 & 15 & 86 \\ 
         \specialrule{0.75pt}{0.5pt}{0.5pt}
         Total & 87,252 & 96,679 \\
         \specialrule{1pt}{0.5pt}{0.5pt}
    \end{tabular}
\end{table}

\begin{table}[h!]
    \caption{Frequency distribution of rationale length in word-level tokens of human-created rationales associated with SEER cancer pathology reports. The class-wise proportion of the training rationales for the 10 common primary cancer sites shows a breakdown in rationale length. The maximum length of a rationale is 128 words.}
    \vspace{1mm}
    \label{tab:rationale_length}
    \resizebox{1\linewidth}{!}{%
    \begin{tabular}{c c c c c c c c c c c c c}
         \specialrule{1pt}{0.5pt}{0.5pt}
         & \multicolumn{2}{c}{Split} & \multicolumn{10}{c}{Class-Wise Training Rationale Proportions in \% } \\
         \cmidrule(lr){2-3} \cmidrule(lr){4-13}
         \multirow{-2}{*}{\shortstack{Rationale Length Range\\in Word Tokens}} & Train & Test 
         & C16 & C18 & C25 & C42 & C44 
         & C50 & C54 & C61 & C64 & C73 \\
         \specialrule{1pt}{0.5pt}{0.5pt}
         1-2 & 16,220 & 452 
         & 23.3
         & 16.6
         & 18.2
         & 24.5
         & 12.9
         & 13.4        
         & 26.5
         & 50.4
         & 15.3
         & 21.6 \\
         3-5 & 17,322 & 429
         & 16.5
         & 15.1
         & 16.3
         & 17.8
         & 28.3
         & 15.0        
         & 21.4
         & 13.8
         & 13.8
         & 24.0 \\
         6-10 & 16,812 & 433
         & 25.1
         & 23.8
         & 19.3
         & 13.7
         & 18.9
         & 14.3        
         & 16.8
         & 12.0
         & 23.2
         & 15.6 \\
         11-15 & 16,503 & 452
         & 14.6
         & 15.3
         & 18.6
         & 14.0
         & 14.5
         & 18.5        
         & 11.7
         & 8.5
         & 21.4
         & 14.0 \\
         16-25 & 14,185 & 359
         & 9.1
         & 10.7
         & 13.6
         & 14.6
         & 10.0
         & 18.4        
         & 9.9
         & 4.8
         & 12.3
         & 11.2 \\
         26+ & 15,637 & 321
         & 11.3
         & 18.6
         & 14.1
         & 15.4 
         & 14.3
         & 20.5        
         & 13.7
         & 10.5
         & 14.0
         & 13.5 \\
         \specialrule{0.75pt}{0.5pt}{0.5pt}
         Total & 96,679 & 2,446 \\
        \specialrule{1pt}{0.75pt}{0.75pt}
        \multicolumn{13}{p{\linewidth}}{
        Ten common primary cancer sites: stomach (C16), colon (C18), pancreas (C25), hematopoietic and
        reticuloendothelial systems (C42), skin (C44), breast (C50), corpus uteri (C54), prostate gland (C61), kidney (C64), and thyroid gland~(C73).} \\
        \specialrule{1pt}{0.75pt}{0.75pt}
    \end{tabular}
    }
\end{table}

\begin{table}[p]
    \centering
        \caption{Frequency distribution of primary cancer sites across training, validation, and test datasets from SEER cancer registries in New Jersey, Louisiana, and Utah. For comparison, the population of available reports annotated with rationales for the primary cancer site and the entire population of reports used to populated the validation and test splits.}
    \label{tab:freq_counts}

\resizebox{1\linewidth}{!}{%
\addtolength{\tabcolsep}{-0.2em}
\begin{tabular}{lcccccccccc}
\toprule
& \multicolumn{2}{c}{Training $\uparrow$} 
& \multicolumn{2}{c}{Validation} 
& \multicolumn{2}{c}{Test} 
& \multicolumn{2}{c}{\makecell[c]{Pop. of Reports\\with Rationales}} 
& \multicolumn{2}{c}{\makecell[c]{Total population\\of Reports}} \\
& \multicolumn{2}{c}{n=87,252} 
& \multicolumn{2}{c}{n=19,385} 
& \multicolumn{2}{c}{n=22,012} 
& \multicolumn{2}{c}{n=172,120} 
& \multicolumn{2}{c}{n=1,572,346} \\
\multirow{-4}{*}{\makecell[c]{Primary Cancer Site\\(ICD-O-3 Code)}}
& Count &(\%) 
& Count &(\%)
& Count &(\%) 
& Count &(\%) 
& Count &(\%) \\

\midrule
Breast (C50) & 25,160 & (28.836) & 5,051 & (26.056) & 5,742 & (26.086) & 50,688 & (29.449) & 403,405 & (25.656) \\
Bronchus and lung (C34) & 10,862 & (12.449) & 1,844 & (9.513) & 2,250 & (10.222) & 19,043 & (11.064) & 153,856 & (9.785) \\
\makecell[l]{Hematopoietic and reticulo-\\endothelial systems (C42)}  & 7,636 & (8.752) & 1,577 & (8.135) & 1,715 & (7.791) & 8,851 & (5.142) & 124,516 & (7.919) \\
Skin (C44) & 5,646 & (6.471) & 1,294 & (6.675) & 1,419 & (6.446) & 9,018 & (5.239) & 101,049 & (6.427) \\
Lymph nodes (C77) & 5,225 & (5.988) & 864 & (4.457) & 1,006 & (4.570) & 13,914 & (8.084) & 67,875 & (4.317) \\
Colon (C18) & 3,679 & (4.217) & 1,009 & (5.205) & 1,144 & (5.197) & 7,834 & (4.551) & 85,430 & (5.433) \\
Corpus uteri (C54) & 3,521 & (4.035) & 679 & (3.503) & 798 & (3.625) & 4,722 & (2.743) & 54,854 & (3.489) \\
Bladder (C67) & 3,306 & (3.789) & 776 & (4.003) & 927 & (4.211) & 4,227 & (2.456) & 66,271 & (4.215) \\
Pancrease (C25) & 1,849 & (2.119) & 327 & (1.687) & 466 & (2.117) & 3,222 & (1.872) & 27,910 & (1.775) \\
Prostate gland (C61) & 1,847 & (2.117) & 1,601 & (8.259) & 1,585 & (7.201) & 2,665 & (1.548) & 128,659 & (8.183) \\
Rectum (C20) & 1,596 & (1.829) & 391 & (2.017) & 348 & (1.581) & 2,324 & (1.350) & 29,489 & (1.875) \\
Stomach (C16) & 1,612 & (1.848) & 259 & (1.336) & 429 & (1.949) & 2,725 & (1.583) & 26,740 & (1.701) \\
Thyroid gland (C73) & 1,500 & (1.719) & 405 & (2.089) & 529 & (2.403) & 1,889 & (1.097) & 33,750 & (2.146) \\
Ovary (C56) & 1,106 & (1.268) & 245 & (1.264) & 228 & (1.036) & 1,482 & (0.861) & 18,617 & (1.184) \\
Kidney (C64)& 1,068 & (1.224) & 376 & (1.940) & 409 & (1.858) & 1,556 & (0.904) & 27,936 & (1.777) \\
Brain (C71) & 980 & (1.123) & 219 & (1.130) & 212 & (0.963) & 1,925 & (1.118) & 18,116 & (1.152) \\
Esophagus (C15) & 826 & (0.947) & 112 & (0.578) & 180 & (0.818) & 1,712 & (0.995) & 13,140 & (0.836) \\
\makecell[l]{Unknown primary site (C80)}  & 744 & (0.853) & 198 & (1.021) & 164 & (0.745) & 15,593 & (9.059) & 12,509 & (0.796) \\
Cervix uteri (C53) & 620 & (0.711) & 167 & (0.861) & 137 & (0.622) & 911 & (0.529) & 11,301 & (0.719) \\
\makecell[l]{Liver and intrahepatic\\bile ducts (C22)}  & 582 & (0.667) & 124 & (0.640) & 187 & (0.850) & 838 & (0.487) & 12,609 & (0.802) \\
\makecell[l]{Connective, subcutaneous\\and other soft tissues (C49)}  & 536 & (0.614) & 125 & (0.645) & 174 & (0.790) & 1,319 & (0.766) & 11,804 & (0.751) \\
Rectosigmoid junction (C19) & 450 & (0.516) & 119 & (0.614) & 123 & (0.559) & 1,486 & (0.863) & 9,168 & (0.583) \\
Small intestine (C17) & 459 & (0.526) & 101 & (0.521) & 153 & (0.695) & 703 & (0.408) & 8,651 & (0.550) \\
Vulva (C51) & 455 & (0.521) & 84 & (0.433) & 93 & (0.422) & 685 & (0.398) & 9,330 & (0.593) \\
Larynx (C32) & 382 & (0.438) & 121 & (0.624) & 113 & (0.513) & 693 & (0.403) & 9,726 & (0.619) \\
\makecell[l]{Other and unspecified\\parts of tongue (C02)}  & 343 & (0.393) & 57 & (0.294) & 78 & (0.354) & 757 & (0.440) & 5,861 & (0.373) \\
Tonsil (C09) & 333 & (0.382) & 76 & (0.392) & 112 & (0.509) & 370 & (0.215) & 7,501 & (0.477) \\
Anus and anal canal (C21) & 327 & (0.375) & 66 & (0.340) & 93 & (0.422) & 618 & (0.359) & 5,491 & (0.349) \\
\makecell[l]{Other and unspecified\\female genital organs (C57)}  & 276 & (0.316) & 53 & (0.273) & 46 & (0.209) & 1,555 & (0.903) & 3,161 & (0.201) \\
Meninges (C70) & 303 & (0.347) & 56 & (0.289) & 81 & (0.368) & 408 & (0.237) & 5,596 & (0.356) \\
Base of tongue (C01) & 272 & (0.312) & 74 & (0.382) & 59 & (0.268) & 248 & (0.144) & 6,143 & (0.391) \\
Renal pelvis (C65) & 253 & (0.290) & 35 & (0.181) & 88 & (0.400) & 245 & (0.142) & 4,631 & (0.295) \\
\makecell[l]{Other and unspecified\\parts of biliary tract (C24)}  & 257 & (0.295) & 58 & (0.299) & 65 & (0.295) & 567 & (0.329) & 4,066 & (0.259) \\
Parotid gland (C07) & 222 & (0.254) & 48 & (0.248) & 53 & (0.241) & 247 & (0.144) & 3,212 & (0.204) \\
\makecell[l]{Heart, mediastinum, and\\pleura (C38)} & 203 & (0.233) & 36 & (0.186) & 58 & (0.263) & 424 & (0.246) & 3,663 & (0.233) \\
Ureter (C66) & 200 & (0.229) & 29 & (0.150) & 37 & (0.168) & 261 & (0.152) & 3,046 & (0.194) \\
\makecell[l]{Retroperitoneum\\and peritoneum (C48)}  & 177 & (0.203) & 44 & (0.227) & 53 & (0.241) & 370 & (0.215) & 3,576 & (0.227) \\
\makecell[l]{Other and ill-defined\\digestive organs (C26)}  & 153 & (0.175) & 19 & (0.098) & 30 & (0.136) & 970 & (0.564) & 1,960 & (0.125) \\
Oropharynx (C10) & 159 & (0.182) & 16 & (0.083) & 23 & (0.104) & 189 & (0.110) & 1,802 & (0.115) \\
Gallbladder (C23) & 146 & (0.167) & 47 & (0.242) & 51 & (0.232) & 147 & (0.085) & 3,089 & (0.196) \\
\makecell[l]{Other endocrine glands\\and related structures (C75)}  & 157 & (0.180) & 60 & (0.310) & 62 & (0.282) & 225 & (0.131) & 4,000 & (0.254) \\
\makecell[l]{Other and unspecified\\parts of mouth (C06)}  & 147 & (0.168) & 42 & (0.217) & 44 & (0.200) & 321 & (0.186) & 2,401 & (0.153) \\
Vagina (C52) &135 & (0.155) & 14 & (0.072) & 15 & (0.068) & 364 & (0.211) & 1,751 & (0.111) \\

\bottomrule
    \end{tabular}

}
\end{table}

\begin{table}[p]
    \centering
\caption*{{Table A.4 (cont.): Frequency distribution of primary cancer sites across training, validation, and test datasets from SEER cancer registries in New Jersey, Louisiana, and Utah. For comparison, the population of available reports annotated with rationales for the primary cancer site and the entire population of reports used to populated the validation and test splits.}}
\label{tab:freq_counts_cont}
\resizebox{1\linewidth}{!}{%
\addtolength{\tabcolsep}{-0.2em}
    \begin{tabular}{lccccccccccc}
\toprule

& \multicolumn{2}{c}{Training $\uparrow$} 
& \multicolumn{2}{c}{Validation} 
& \multicolumn{2}{c}{Test} 
& \multicolumn{2}{c}{\makecell[c]{Pop. of Reports\\with Rationales}} 
& \multicolumn{2}{c}{\makecell[c]{Total population\\of Reports}} \\
& \multicolumn{2}{c}{n=87,252} 
& \multicolumn{2}{c}{n=19,385} 
& \multicolumn{2}{c}{n=22,012} 
& \multicolumn{2}{c}{n=172,120} 
& \multicolumn{2}{c}{n=1,572,346} \\
\multirow{-4}{*}{\makecell[c]{Primary Cancer Site\\(ICD-O-3 Code)}}
& Count & (\%) 
& Count & (\%)
& Count & (\%) 
& Count & (\%) 
& Count & (\%) \\

\midrule
Testis (C62) & 134 & (0.154) & 67 & (0.346) & 81 & (0.368) & 301 & (0.175) & 4,889 & (0.311) \\
\makecell[l]{Bones, joints, and articular\\cartilage of other and\\unspecified sites (C41)} & 115 & (0.132) & 30 & (0.155) & 35 & (0.159) & 308 & (0.179) & 2,530 & (0.161) \\
Nasal cavity and middle ear (C30) & 111 & (0.127) & 22 & (0.113) & 29 & (0.132) & 169 & (0.098) & 1,498 & (0.095) \\
Uterus, NOS (C55) & 101 & (0.116) & 25 & (0.129) & 17 & (0.077) & 284 & (0.165) & 1,534 & (0.098) \\
\makecell[l]{Spinal cord, cranial nerves,\\and other parts of central\\nervous system (C72)} & 107 & (0.123) & 36 & (0.186) & 26 & (0.118) & 137 & (0.080) & 2,239 & (0.142) \\
\makecell[l]{Other and unspecified\\urinary organs (C68)} & 98 & (0.112) & 23 & (0.119) & 20 & (0.091) & 677 & (0.393) & 1,710 & (0.109) \\
Nasopharynx (C11) & 93 & (0.107) & 28 & (0.144) & 31 & (0.141) & 112 & (0.065) & 1,740 & (0.111) \\
Penis (C60) & 82 & (0.094) & 11 & (0.057) & 12 & (0.055) & 142 & (0.083) & 1,596 & (0.102) \\
\makecell[l]{Bones, joints and articular\\cartilage of limbs (C40)} & 76 & (0.087) & 32 & (0.165) & 21 & (0.095) & 108 & (0.063) & 1,990 & (0.127) \\
Accessory sinuses (C31) & 67 & (0.077) & <10 & (0.031) & 14 & (0.064) & 70 & (0.041) & 1,192 & (0.076) \\
Eye and adnexa (C69) & 63 & (0.072) & <10 & (0.005) & 13 & (0.059) & 108 & (0.063) & 1,620 & (0.103) \\
Thymus (C37) & 65 & (0.074) & <10 & (0.036) & <10 &(0.014) & 98 & (0.057) & 773 & (0.049) \\
Palate (C05) & 56 & (0.064) & 48 & (0.248) & 18 & (0.082) & 136 & (0.079) & 1,487 & (0.095) \\
Floor of mouth (C04) & 54 & (0.062) & 18 & (0.093) & 12 & (0.055) & 112 & (0.065) & 1,727 & (0.110) \\
Other and ill-defined sites (C76) & 51 & (0.058) & <10 & (0.026) & <10 & (0.018) & 363 & (0.211) & 766 & (0.049) \\
\makecell[l]{Other and unspecified\\major salivary glands (C08)} & 51 & (0.058) & 28 & (0.144) & <10 & (0.032) & 115 & (0.067) & 910 & (0.058) \\
Hypopharynx (C13) & 47 & (0.054) & 11 & (0.057) & <10 & (0.023) & 56 & (0.033) & 985 & (0.063) \\
Gum (C03) & 41 & (0.047) & 25 & (0.129) & 15 & (0.068) & 103 & (0.060) & 1,273 & (0.081) \\
Lip (C00) & 23 & (0.026) & 10 & (0.052) & <10 & (0.027) & 109 & (0.063) & 926 & (0.059) \\
\makecell[l]{Other and ill-defined sites in\\lip, oral cavity and pharynx (C14)} & 24 & (0.028) & <10 & (0.046) & 31 & (0.141) & 112 & (0.065) & 702 & (0.045) \\
Adrenal gland (C74) & 26 & (0.030) & 12 & (0.062) & 13 & (0.059) & 83 & (0.048) & 1,008 & (0.064) \\
Pyriform sinus (C12) & 26 & (0.030) & <10 & (0.041) & 11 & (0.050) & 31 & (0.018) & 624 & (0.040) \\
\makecell[l]{Peripheral nerves and\\autonomic nervous system (C47)} & 11 & (0.013) & 25 & (0.129) & <10 & (0.014) & 11 & (0.006) & 392 & (0.025) \\
\makecell[l]{Other and unspecified\\male genital organs (C63)} & 12 & (0.014) & 0 & (0.000) & <10 & (0.014) & 29 & (0.017) & 338 & (0.021) \\
Placenta (C58) & <10 & (0.006) & 0 & (0.000) & 0 & (0.000) & 0 & (0.000) & 55 & (0.003) \\
Trachea (C33) & <10 & (0.003) & 0 & (0.000) & <10 & (0.014) & 35 & (0.020) & 181 & (0.012) \\
\bottomrule
    \end{tabular}
}

\end{table}

\clearpage

\subsection{Model Architectures} \label{app:mm:model_architecture}

Here, we briefly describe the evaluated model architectures. We selected the clinical longformer (CLF) with its baseline weights (CLF-BS) as the text-encoder architecture for all models because the CLF has enabled models to achieve exceptional performance on clinical text classification tasks \cite{li2022clinical}. Our specific implementation is from the HuggingFace transformer library. The CLF utilizes a byte-pair encoded vocabulary to tokenize the incoming text sequences. A pretrained subword embedding layer is used to transform each subword token into a vector representation with a hidden dimension of $d_h = 768$. The CLF-BS is the baseline model for all experiments. The second evaluated model utilizes a recently proposed multiple attention mechanism to learn lexical and contextual information from the input sequence relevant to our classification task \cite{metzner2024deformable}. %The following paragraph describes the classification process performed by our models in more detail.

Given an input text sequence $X=[x_1, x_2, \ldots, x_N]$ with $N$ tokens, each token is mapped to its corresponding embedding vector representation with size $d_e$, creating a token-embedding matrix $X_E \in \mathbb{R}^{N \times d_e}$. $X_E$ is then processed by a text-encoder architecture (i.e., CLF) to learn a latent document matrix $H \in \mathbb{R}^{N \times d_h}$, where $d_h$ is the hidden dimension of the underlying architecture. Depending on the evaluated model architecture---CLF-BS or CLF with a deformable phrase-level attention mechanism (CLF-DPLA)---the latent document matrix $H$ is either used directly as the final document context vector representation $C \in \mathbb{R}^{1 \times d_h}$ or further processed. For the CLF-BS, we follow standard practice with transformers and used the hidden states $H_0 \in \mathbb{R}^{1 \times d_h}$ of the first special token $<\!\!s\!\!>$ as $C$. In contrast, the CLF-DPLA deploys a complex attention mechanism to further encode $H$ to enrich $C$ with lexical word-level and contextual phrase-level information relevant to the classification task. $C$ is then passed to a linear classification layer (\autoref{eq:classification_layer}) with projection weights $W \in \mathbb{R}^{|L| \times d_h}$ and bias $B \in \mathbb{R}^{|L| \times 1}$. $|L|$ is the number of classes for a given task. The output of the classification layer is then passed to a softmax activation function, thereby creating class-wise pseudo-probabilities. The class with the maximum probability is selected as the prediction $\hat{y}$ for a given text sequence. We refer the reader to \cite{metzner2024deformable} for more details. All text sequences were tokenized by utilizing the byte-pair encoded vocabulary associated with the CLF, and the maximum sequence length was 4,096 tokens.

\begin{equation} \label{eq:classification_layer}
    \hat{y}(C, W, B) = \arg\max_{y} softmax(C \cdot W^\top+ B)
\end{equation}

\subsection{Model Training} \label{app:model_training}

Starting from the pretrained weights of the CLF-BS, we fine-tune our CLF models to convergence  \cite{li2022clinical}. The model weights are optimized by using the AdamW optimizer and the cross-entropy objective loss function. Training is accelerated by using automatic mixed precision and is parallelized across two compute nodes of the Frontier supercomputer via the CITADEL security framework. Because SEER electronic pathology reports contain protected health information, we utilized the CITADEL security framework associated with the Scalable Protected Infrastructure provided by the National Center for Computational Sciences and the Oak Ridge Leadership Computing Facility. Each Frontier compute node is equipped with four AMD MI250X GPUs (with two Graphics Compute Dies in each GPU), which are equal to eight 64~GB memory GPUs. To stabilize the fine-tuning process, we employed a linear learning rate scheduler for the first five epochs and implemented early stopping after three epochs to prevent overfitting. We utilized the hyperparameters identified in \cite{metzner2024deformable} and set the batch size to eight.

\subsection{Model Evaluation} % DONE

We evaluate the performance of all models using quantitative metrics established in the multiclass text classification literature and measured the performance accuracy (\autoref{eq:accuracy}) and macro-averaged F1-score (\autoref{eq:macro-F1}) \cite{kowsari2019text, gao2021limitations, minaee2021deep}. We use the widely accepted normal approximation \cite{raschka2018model} to compute 95\% confidence intervals. 

\begin{equation}\label{eq:accuracy}
    \text{Accuracy} = \frac{TP+TN}{TP+FP+FN+TN}
\end{equation}
\begin{equation} \label{eq:macro-F1}
    \text{Macro-F1-Score} = \frac{1}{C} \sum_{c=i}^{C} F1\textnormal{-}Score_c
\end{equation}
\begin{equation}\label{eq:classwise_metrics}
    \text{Precision}_c = \frac{TP_c}{TP_c+FP_c} \\
\end{equation}
\begin{equation*}
    \text{Recall}_c = \frac{TP_c}{TP_c+FN_c} \\
\end{equation*}
\begin{equation*}
    \text{F1-Score}_c = 2\times\frac{Precision_c\times Recall_c}{Precision_c + Recall_c},\\   
\end{equation*}

\noindent where $C$ is the total number of possible classes in the task-specific label space, $c$ is the $i^{th}$ class, $TP$ is true positive, $TN$ is true negative, $FP$ is false positive, and $FN$ is false negative. In the multi-class classification setting, accuracy computes the global true positives, false positives, false negatives, and true negatives by treating all classes as binary classification and then sums up all the class-wise counts. To calculate the class-wise $\text{F1-score}_c$, we treat the $TP_c$, $FP_c$, and $FN_c$ scores as the true positives, false positives, and false negatives, respectively, when evaluating class c against all other classes.

\clearpage

\section{Results} \label{app:results}
\setcounter{table}{0}
\setcounter{equation}{0}
\counterwithin{figure}{section}
\setcounter{figure}{0}
\setcounter{page}{1}

\subsection{Preliminary Results} \label{app:res:preliminary_results}

During the preliminary phase of the study, we experimented with various algorithmic approaches to integrate the rationale information into the models. Primarily, we experimented with approaches that integrated rationale information as binary masks, which indicate whether or not a token is a part of a rationale. The first model utilizes a CLF-BS as the text encoder architecture to encode the input sequence and two multitask classification layers to perform token-sequence classification.
The first layer performs token classification by trying to predict whether a token is part of a rationale. It does this by using the hidden states of all tokens of the sequences as input and the binary rationale masks as the ground-truth labels. The second layer utilizes the hidden states of the first special token as the document vector representation to perform its prediction for the primary cancer site.
The loss from the token classification is scaled with $\alpha = 0.25$ and added to the full loss from the sequence classification. We evaluated $\alpha \in [0.25, 0.5, 0.75]$ to optimize the models' performance. 
The second model we investigated extends the CLF-BS with a masked multihead self-attention layer using the binary rationale mask to control the attention flow of the two self-attention heads. We optimized for the number of heads from $m_{heads}=[2, 4, 8]$. Conceptually, this model is designed to optimize its parameters to focus on information that is contextually similar to the rationales.

\autoref{tab:intro_preliminary_results} overviews the preliminary experiments performed to determine the optimal method for incorporating human-based clinical rationales into the training procedure of transformer-based clinical text classification models. We focused our experiments on the CLF because it has shown favorable performance in clinical text classification tasks \cite{metzner2024attention}. The evaluated methods include data augmentation (i.e., using the rationales as additional samples in the training dataset) and changes to the underlying algorithm following proposed models in the literature. For the algorithmic approach, we explored the use of binary masks to indicate whether a token is part of a rationale as an additional input to a multitask, learning-based token-sequence classification model. These masks functioned as the ground-truth labels for token classification and as an attention mask used in a masked multihead self-attention head. The preliminary experiments show that the addition of rationales to the training dataset outperformed the algorithmic implementations, and this is why we selected the rationales as the primary approach to incorporating human expert--derived input in ML model training.

\begin{table}[h!]
    \centering
    \caption{Preliminary results for incorporating human-based rationales in the CLF.}
    \vspace{1mm}
    \begin{tabular}{l l c c}
        \specialrule{1.0pt}{0pt}{0pt}
        Clinical Longformer  & Training Data Input  & F1-Macro & Accuracy \\
        \specialrule{1.0pt}{0pt}{0pt}
        Baseline & Reports only & 0.547 & 0.881 \\
        + Deformable Phrase-Level Attention & Reports only & 0.582 & 0.898 \\
        \specialrule{0.75pt}{0pt}{0pt}
        Baseline & Reports and Rationales & 0.562 & 0.885 \\
        + Deformable Phrase-Level Attention & Reports and Rationales  & \textbf{0.592} & \textbf{0.902} \\
        \specialrule{0.75pt}{0pt}{0pt}
        + Token-Sequence-Classification & Reports and Rationale Masks & 0.525 & 0.881 \\
        + Masked Multi-Head Self-Attention & Reports and Rationale Masks & 0.573 & 0.897 \\
        \specialrule{1.0pt}{0pt}{0pt}
        \multicolumn{4}{l}{Dataset sizes: $n_{train}=90,660$, $n_{train, rationales}=100,204$, $n_{val}=19390$, and $n_{test}=19,522$)}\\
        \specialrule{1pt}{0pt}{0pt}
    \end{tabular}
    \label{tab:intro_preliminary_results}
\end{table}

\subsection{Results: Performance}
\begin{table}[h!]
    \centering
    \caption{Improvements in accuracy and F1-macro scores with the addition of $m$ randomly selected, \textit{sufficient}, human-generated rationales associated with 10 common primary cancer sites for the CLF-BS and CLF-DPLA models when classifying primary cancer sites from SEER cancer pathology reports. The $\uparrow$ and $\downarrow$ indicate increased or decreased performance, respectively, compared to the model trained with only reports. Scores marked with -- have equal performance.}
    \vspace{1mm}
    \resizebox{1\linewidth}{!}{%
    \begin{tabular}{c c c c c c}
    \specialrule{1pt}{0.5pt}{0.5pt}
     & & & \multicolumn{3}{c}{Sufficiency Threshold}\\
     \cmidrule(lr){4-6}
     
    \multirow{-2}{*}{Model} & \multirow{-2}{*}{\shortstack{Number of Added\\Rationales}} & \multirow{-2}{*}{Metric} & > 0 & > 0.2 & > 0.817\\
    \specialrule{1pt}{0.5pt}{0.5pt}

     &  & \multirow{-2}{*}{Accuracy} 
     & \shortstack{0.884$\downarrow$\\(0.880, 0.888)} 
     & \shortstack{0.884$\downarrow$\\(0.880, 0.888)}
     & \shortstack{0.885 -- \\(0.880, 0.889)}\\
     & \multirow{-3}{*}{100} 
     & \multirow{-2}{*}{F1-Macro} 
     & \shortstack{0.555$\downarrow$\\(0.549, 0.562)} 
     & \shortstack{0.565$\downarrow$\\(0.558, 0.572)}
     & \shortstack{0.564$\downarrow$\\(0.557, 0.571)}\\
     \cmidrule(lr){2-6}
     &  & \multirow{-2}{*}{Accuracy} 
     &  \shortstack{0.886$\uparrow$\\(0.881, 0.890)} 
     &  \shortstack{0.882$\downarrow$\\(0.878, 0.887)}
     & \shortstack{0.884$\downarrow$\\(0.880, 0.889)}\\
     \multirow{-3}{*}{CLF-BS}
     & \multirow{-3}{*}{500} 
     & \multirow{-2}{*}{F1-Macro}  
     & \shortstack{0.568$\uparrow$\\(0.561, 0.574)} 
     & \shortstack{0.563$\downarrow$\\(0.556, 0.569)}
     & \shortstack{0.565$\downarrow$\\(0.558, 0.571)}\\
     \cmidrule(lr){2-6}
     & & \multirow{-2}{*}{Accuracy} 
     & \shortstack{0.884$\downarrow$\\(0.880, 0.889)} 
     & \shortstack{0.885 -- \\(0.880, 0.889)}
     & \shortstack{0.884$\downarrow$\\(0.884, 0.884)}\\
     & \multirow{-3}{*}{1,000} 
     & \multirow{-2}{*}{F1-Macro}  
     & \shortstack{0.570$\uparrow$\\(0.563, 0.577)} 
     & \shortstack{0.565$\downarrow$\\(0.558, 0.571)}
     & \shortstack{0.570$\uparrow$\\(0.563, 0.577)}\\
    \specialrule{1pt}{0.5pt}{0.5pt}
    &  & \multirow{-2}{*}{Accuracy} 
    &\shortstack{0.898$\downarrow$\\(0.894, 0.902)} 
    &  \shortstack{0.899$\downarrow$\\(0.895, 0.903)} 
    & \shortstack{0.900 -- \\(0.896,v0.904)}\\
     & \multirow{-3}{*}{100} 
     & \multirow{-2}{*}{F1-Macro} 
     & \shortstack{0.583$\downarrow$\\(0.576, 0.590)}
     & \shortstack{0.582$\downarrow$\\(0.575, 0.589)} 
     & \shortstack{0.584$\downarrow$\\(0.577, 0.591)} \\
     \cmidrule(lr){2-6}
    &  & \multirow{-2}{*}{Accuracy} 
    &\shortstack{0.900 -- \\(0.896, 0.904)} 
    &  \shortstack{0.899$\downarrow$\\(0.895, 0.903)}
    & \shortstack{0.901$\uparrow$\\(0.897, 0.905)}\\
    \multirow{-3}{*}{CLF-DPLA} & \multirow{-3}{*}{500} & \multirow{-2}{*}{F1-Macro} 
    & \shortstack{0.589$\downarrow$\\(0.583, 0.595)}
    & \shortstack{0.596$\downarrow$\\(0.589, 0.602)} 
    & \shortstack{0.580$\downarrow$\\(0.574, 0.587)} \\
     \cmidrule(lr){2-6}

    &  & \multirow{-2}{*}{Accuracy} 
    & \shortstack{0.898$\downarrow$\\(0.894, 0.902)}
    & \shortstack{0.899$\downarrow$\\(0.895, 0.903)}
    & \shortstack{0.897$\downarrow$\\(0.893, 0.901)}\\
    & \multirow{-3}{*}{1,000} & \multirow{-2}{*}{F1-Macro} 
    & \shortstack{0.590$\downarrow$\\(0.583, 0.596)}
    & \shortstack{0.580$\downarrow$\\(0.574, 0.587)}
    &  \shortstack{0.580$\downarrow$\\(0.573, 0.586)} \\
    \specialrule{1pt}{0.5pt}{0.5pt}
    \multicolumn{6}{p{\linewidth}}{%
    Common primary cancer sites: stomach (C16), colon (C18), pancreas (C25),
    hematopoietic and reticuloendothelial systems (C42), skin (C44), breast (C50), corpus uteri (C54),
    prostate gland (C61), kidney (C64), and thyroid gland (C73).
    Sufficiency thresholds: $>$0 includes all rationales, $>$0.2 is the best achievable performance
    improvement, and $>$0.817 represents the 90\% percentile of rationales with false predictions for falsely predicted reports.
    The 95\% confidence interval is computed via normal approximation \cite{raschka2018model}.
    $^a$Utilized all 973 rationales associated with kidney cancer (C64) after removing all non-sufficient rationales.} \\
    \specialrule{1pt}{0.5pt}{0.5pt}
    \end{tabular}
    }
    \label{tab:performance_results_m_suff_rationales}
\end{table}

\clearpage

\subsection{Results: Explainability} \label{app:explainability}
\subsubsection{Average Token-Level Rationale Coverage: Prediction} \label{app:int_prediction}

\begin{figure}[hbt!]
    \centering
    \includegraphics[width=0.8\linewidth]{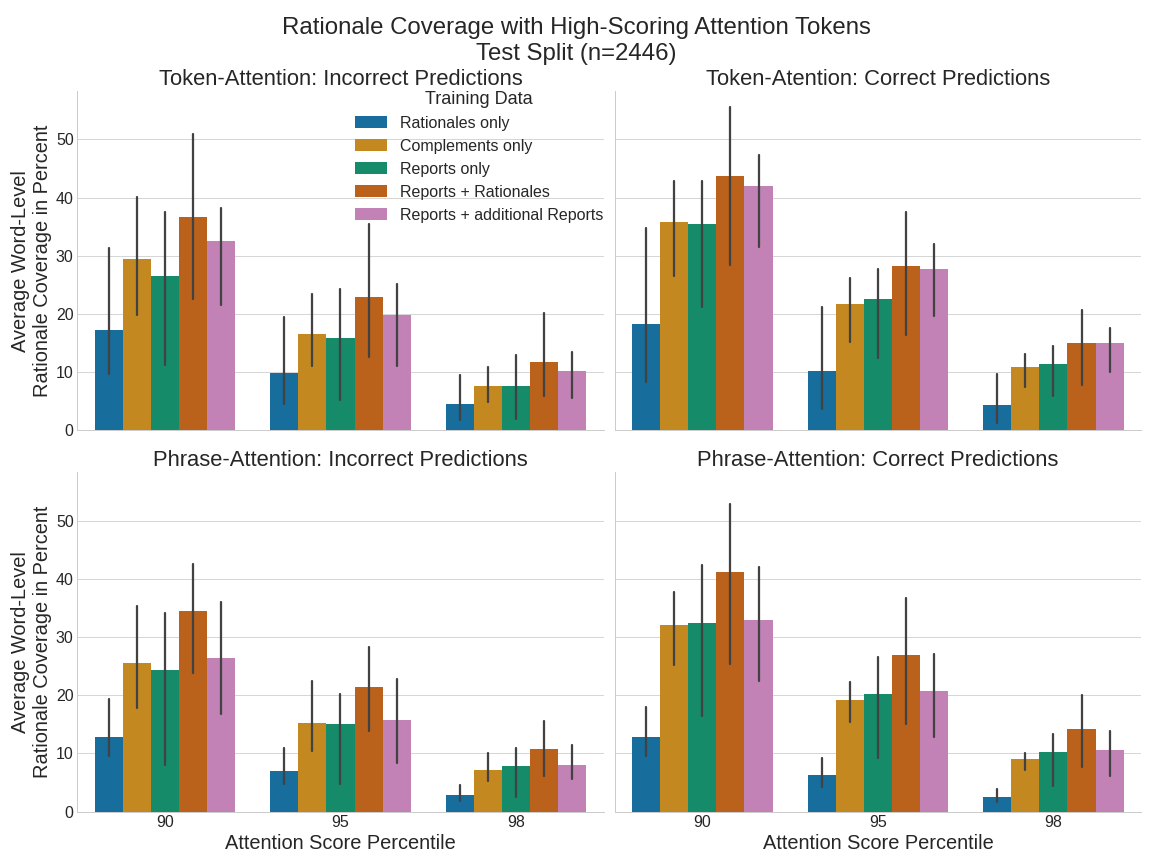}
    \caption{Average word-level rationale coverage broken down by training type and prediction type (correct vs. incorrect) based on attention scores generated by the token- and phrase-level target attention mechanisms of the DPLAs implemented in conjunction with the CLF for different attention score percentiles.}
    \label{fig:rationale_coverage_overlap_2b}
\end{figure}

\subsubsection{Average Token-Level Rationale Coverage: Rationale Length} \label{app:int_length}

\begin{figure}[h!]
    \centering
    \includegraphics[width=0.8\linewidth]{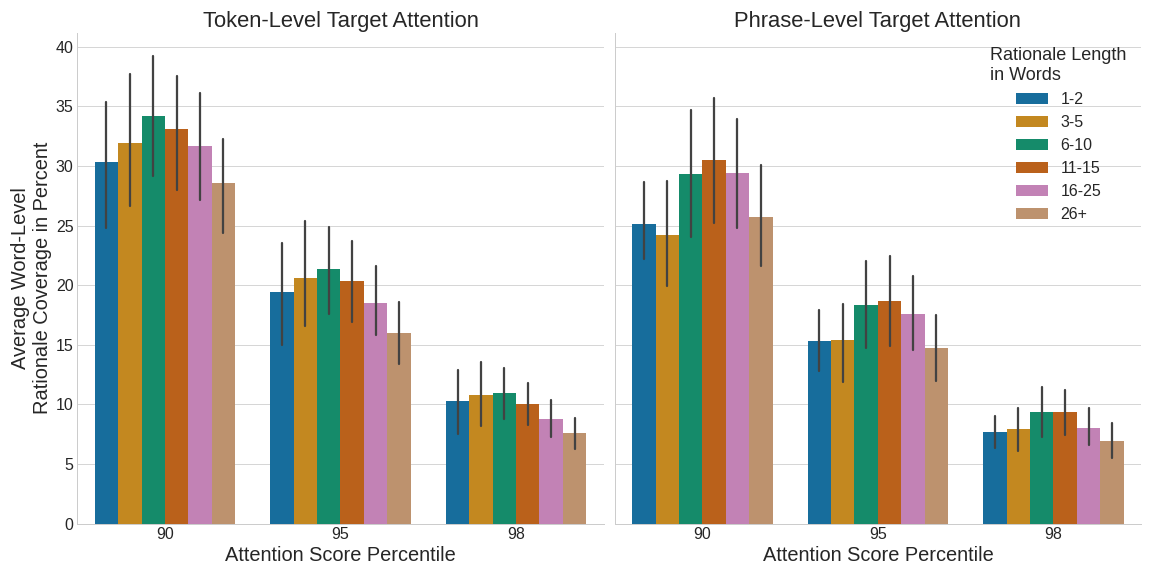}
    \caption{Average word-level rationale coverage broken down by rationale length based on attention scores generated by the token- and phrase-level target attention mechanisms of the DPLAs implemented in conjunction with the CLF for different attention score percentiles.}
    \label{fig:rationale_coverage_overlap_3}
\end{figure}

\clearpage

\subsubsection{Average Token-Level Rationale Coverage: Overlap} \label{app:int_overlap}
The following figures show how models trained on the various regimens highlight up to 60\% of the same tokens associated with the rationales. Each figure represents the view from one training regimen compared with the remaining four training regimens.

\begin{figure}[h!]
  \centering 
  \subfloat[][\centering Overlap of rationale coverage for models trained on reports only.]{\includegraphics[width=1\linewidth]{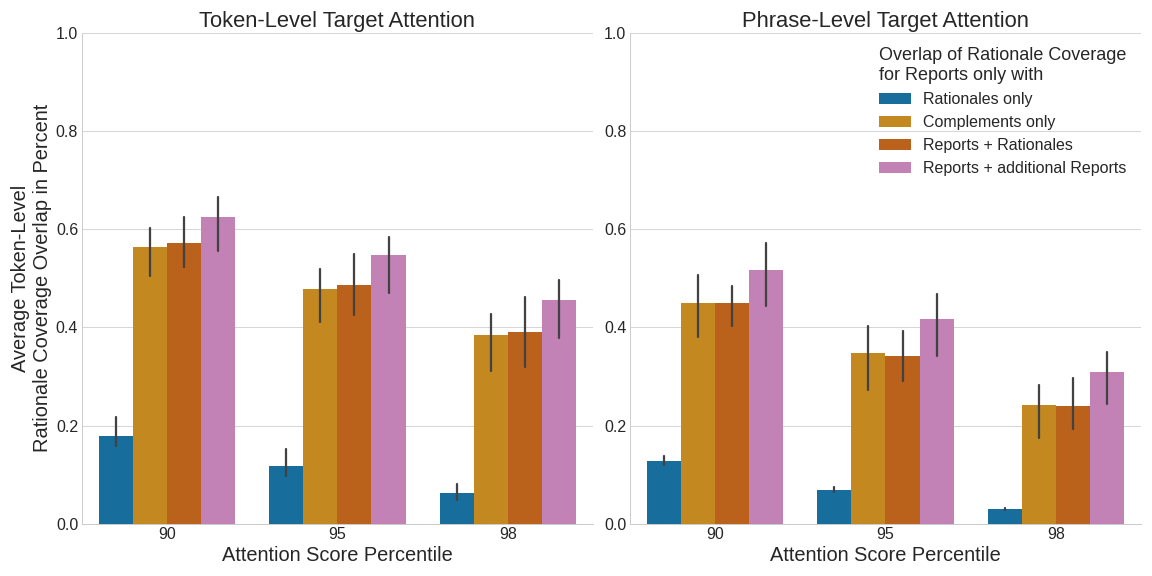}}% 
  \qquad 
  \subfloat[][\centering Overlap of rationale coverage for models trained on rationales only.]{\includegraphics[width=1\linewidth]{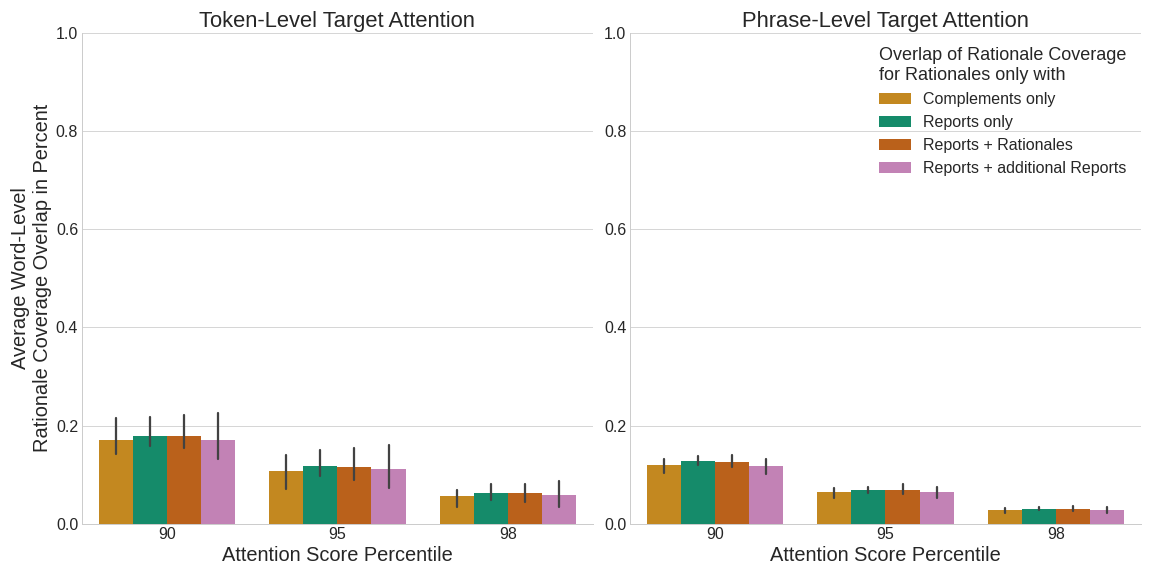}}% 
  \vspace{1mm}
  \caption{Average word-level rationale coverage overlap on test documents associated with clinical rationales for models trained on reports only. Rationale coverage computed based on the attention scores generated by the token- and phrase-level target attention mechanisms of the DPLAs implemented in conjunction with the CLF for different attention score percentiles.}
  \label{fig:cont}
\end{figure}

\begin{figure}[h!]
  \ContinuedFloat 
  \captionsetup{list=off,format=plain}
\captionsetup[subfigure]{format=plain}

  \centering 
  \subfloat[][\centering Overlap of rationale coverage for models trained on complements only.]{\includegraphics[width=1\linewidth]{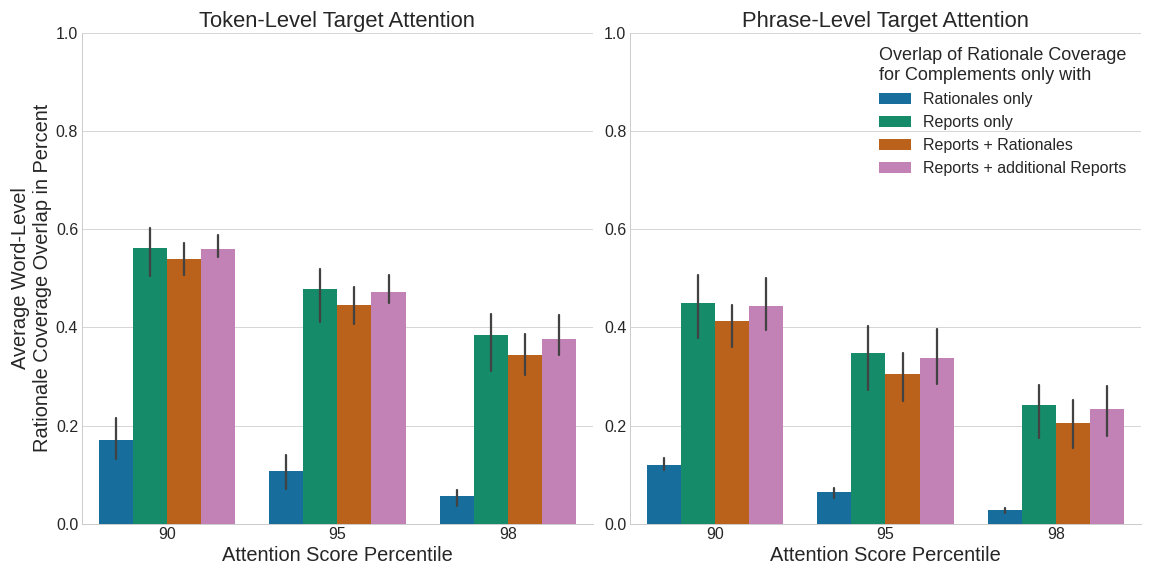}}% 
  \quad 
  \subfloat[][\centering Overlap of rationale coverage for models trained on reports supplemented with human-based clinical rationales.]{\includegraphics[width=1\linewidth]{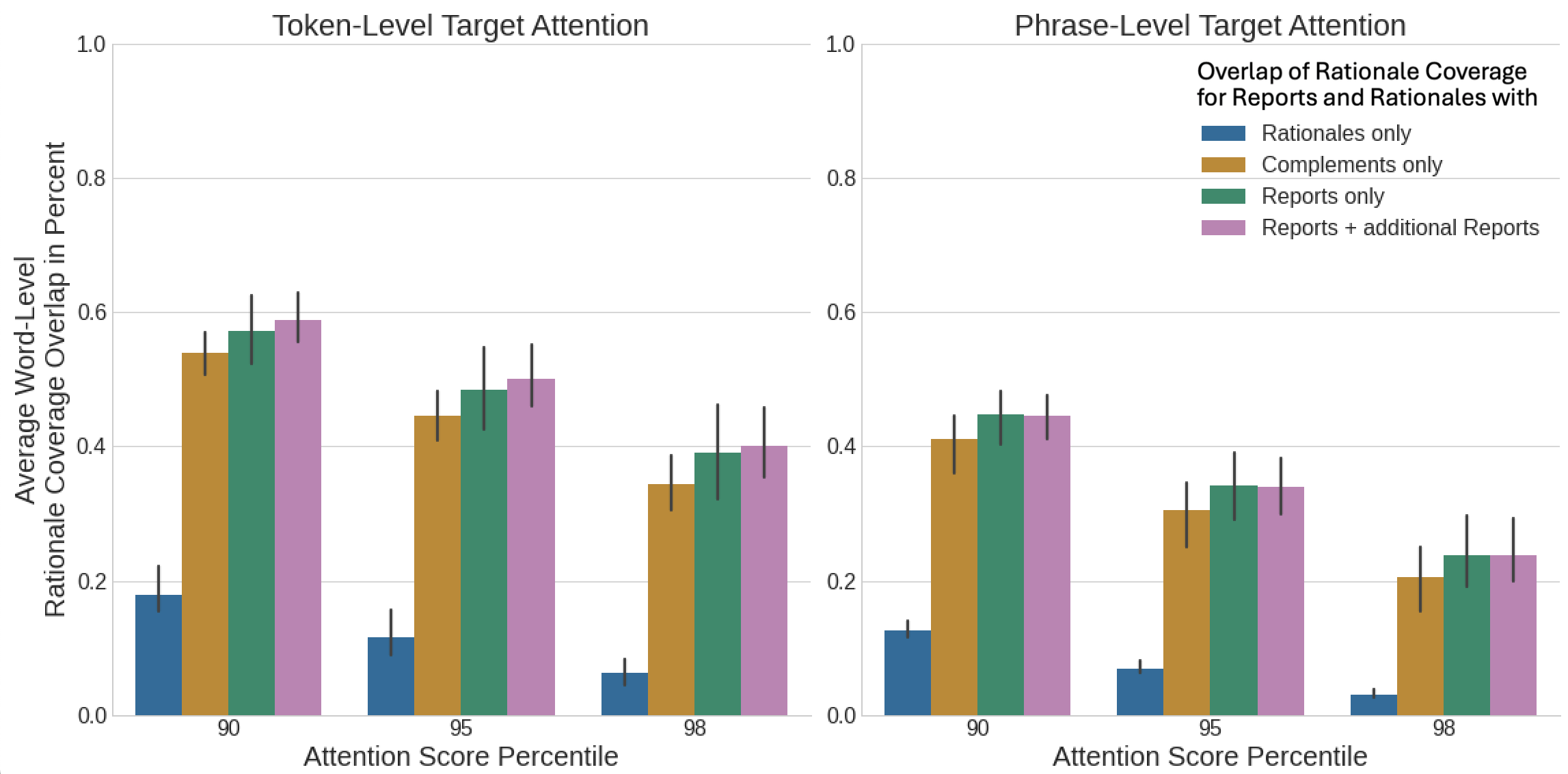}}% 
  \vspace{1mm}
  \caption[]{Overlap of the average token-level rationale coverage for test documents associated with clinical rationales across the five training regimens; complements only and reports supplemented with human-based clinical rationales.}
  \label{fig:cont}
\end{figure} 
 
\begin{figure}[h!]
  \ContinuedFloat 
\captionsetup{list=off,format=plain}
\captionsetup[subfigure]{format=plain}
  \centering 
  \subfloat[][\centering Overlap of rationale coverage for models trained on reports supplemented with additional reports.]{\includegraphics[width=1\linewidth]{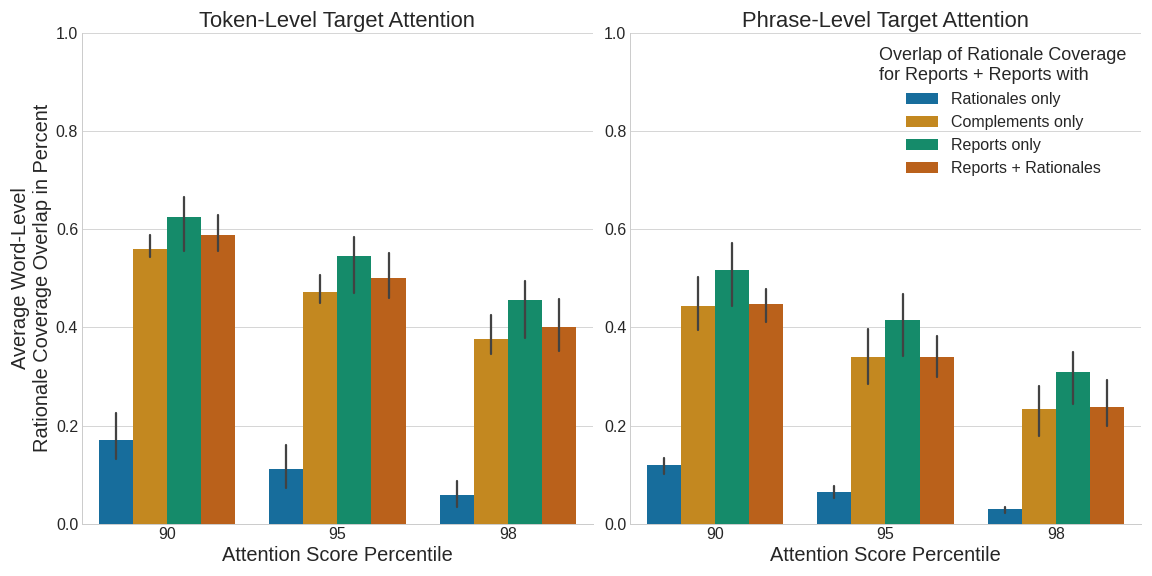}}% 
  \vspace{1mm}
  \caption[]{Overlap of the average token-level rationale coverage for test documents associated with clinical rationales across the five training regimens; reports supplemented with additional electronic pathology reports.}
  \label{fig:cont}
\end{figure} 

\end{document}